\documentclass[preprint,review,11pt,letterpaper]{elsarticle}

\usepackage[margin=1in]{geometry}

\usepackage{graphicx}
\usepackage{algorithm,algorithmic,amsfonts,amsmath,booktabs,multirow,amssymb,bm}
\usepackage{subcaption}
\usepackage{mathtools}
\usepackage{color,soul}
\definecolor{highlight}{rgb}{1.0,1.0,0.0}
\sethlcolor{highlight}

\journal{Pervasive and Mobile Computing}

\begin{document}

\begin{frontmatter}





\title{An Indoor Localization Dataset and Data Collection Framework with High Precision Position Annotation}


\author[1,2]{F. Serhan Dani\c{s} \corref{cor1}}
\author[1]{A. Teoman Naskali}
\author[2]{A. Taylan Cemgil}
\author[2]{Cem Ersoy}
\address[1]{Dept. of Computer Engineering, Galatasaray University, Istanbul, Turkey}
\address[2]{Dept. of Computer Engineering, Bo\u{g}azi\c{c}i University, Istanbul, Turkey}
\cortext[cor1]{Corresponding author: sdanis@gsu.edu.tr}

\begin{abstract}
We introduce a novel technique and an associated high resolution dataset that aims to precisely evaluate wireless signal based indoor positioning algorithms. The technique implements an augmented reality (AR) based positioning system that is used to annotate the wireless signal parameter data samples with high precision position data. We track the position of a practical and low cost navigable setup of cameras and a Bluetooth Low Energy (BLE) beacon in an area decorated with AR markers. We maximize the performance of the AR-based localization by using a redundant number of markers. Video streams captured by the cameras are subjected to a series of marker recognition, subset selection and filtering operations to yield highly precise pose estimations. Our results show that we can reduce the positional error of the AR localization system to a rate under 0.05 meters. The position data are then used to annotate the BLE data that are captured simultaneously by the sensors stationed in the environment, hence, constructing a wireless signal data set with the ground truth, which allows a wireless signal based localization system to be evaluated accurately.
\end{abstract}


\end{frontmatter}
\newcommand{\FIGDIR}{figures/}



\newcommand{\tmat}[2]{\ensuremath{{}^{#1}_{#2}\mathbf{T}}}
\newcommand{\rmat}[2]{\ensuremath{{}^{#1}_{#2}\mathbf{R}}}
\newcommand{\tvec}[2]{\ensuremath{{}^{#1}\boldsymbol{\tau}_{#2}}}
\newcommand{\rvec}[2]{\ensuremath{{}^{#1}\boldsymbol{\theta}_{#2}}}
\newcommand{\markeri}{\ensuremath{m}}
\newcommand{\camera}{\ensuremath{c}}
\newcommand{\world}{\ensuremath{w}}
\newcommand{\obs}{\ensuremath{\mathbf{y}}}
\newcommand{\point}{\ensuremath{\mathbf{x}}}
\newcommand{\error}{\ensuremath{\epsilon}}
\newcommand{\frames}{\ensuremath{F}}
\newcommand{\selection}{\ensuremath{C}}
\newcommand{\outlier}{\ensuremath{U}}
\newcommand{\transM}{\ensuremath{\mathbf{A}}}
\newcommand{\obsM}{\ensuremath{\mathbf{H}}}
\newcommand{\transN}{\ensuremath{\mathbf{R}}}
\newcommand{\obsN}{\ensuremath{\mathbf{Q}}}
\newcommand{\kalmangain}{\ensuremath{\mathbf{K}}}
\newcommand{\predict}{\ensuremath{\mathbf{P}}}
\newcommand{\identity}{\ensuremath{\mathbf{I}}}
\newcommand{\distance}{\ensuremath{d}}
\newcommand{\etal}{\textit{et al.}}

\section{Introduction}


The majority of the world population has started to dwell in cities and urbanized areas, where people spend most of their time indoors. As the buildings and structures become ever more complex, the need to use navigation tools gains importance. However, outdoor navigation technologies like GPS do not work indoors. Therefore, there is a need to use other means of localization indoors.

Wireless technologies are preferred for indoor localization as they are globally available and affordable. Whereas the localization algorithms have been improved with the ubiquitous Wi-Fi technology for indoors \citep{bisio2016b}, due to its low energy consumption, price and size, Bluetooth Low Energy (BLE) is particularly suited to this domain \citep{faragher2015}. BLE beacons or sensors are utilized to track objects or people, and most hand held mobile devices can easily incorporate this technology. However, the nature of the wireless signals render the localization task rather difficult as the signals are easily reflected or absorbed by obstacles. To calibrate and optimize wireless based localization systems, ground truth data must be collected. More importantly, to measure the real performance of these systems, we require trajectories labeled with ground truth positions alongside the wireless signal data, rather than separately measured reference points. This ground truth collection process can be rather tedious and usually labor intensive \citep{torressospedra2019}.

In the literature, there exist several valuable datasets formed by wireless signal parameters and corresponding ground truth information, however, such data collections do not reflect the true nature of the trajectories. Instead, the sets are formed by signal parameters collected with sparsely located reference points \citep{mendozasilva2019}, or by modifying the paths artificially making the carrier remaining still on predefined reference points \citep{baronti2018}. Moreover, the ground truth notion may not even belong to the positions. In the work of Girolami \etal~\citep{girolami2020}, the ground truth of social interactions are reported with the BLE received signal strength indicator (RSSI) data. Likewise, in the work of Sansano-Sansano \etal~\citep{sansanosansano2020}, the ground truth is in terms of the user gate speed.

High resolution ground truth for indoor positioning is rare because of the difficulties in collecting accurate position data on every point the object of interest passes. This difficulty prevents the researchers from setting a generalized benchmark for indoor positioning systems \citep{eyobu2018}. The positioning community lacks a clear standard for evaluating positioning systems. Multiple attempts have been made in order to fill this gap to validate the accuracy of an indoor positioning system. De la Osa \etal~propose a method to define a ground truth for real life purposes \citep{delaosa2016}. The technique bases on creating predefined paths with checkpoints which are used as indicators when the tester passes over them. The ground truth trajectory is then estimated by interpolated points between checkpoints. Another one is the skeleton path technique in which the user follows a predetermined path with a constant speed and the positions are sampled. Clearly the techniques based on walking are prone to human errors, incorrectly assuming constant paces and true timings \citep{poulose2021}. Moreover, the positions are estimated using inaccurate dead-reckoning data based on step counting \citep{ai2020}. According to Adler~\etal~\citep{adler2015}, this predefined path technique, or namely office walk, is the most preferred ground truth collection method. This inaccurate method is also the cheapest and the most precise ground truth data collection method so far used to validate indoor positioning systems.

Our ground truth data collection technique provides a set of timestamped wireless signal parameters. Whereas the technique can be easily modified to collect and annotate other wireless signal based parameters, we work with the prototypical BLE RSSI data. The RSSI data are annotated both in temporal and spatial domains accurately with the corresponding ground truth positions using a vision-based localization system. We employ the markers designed for AR applications, to annotate the RSSI data with ground truth positions.

AR is another technology that performs exceptionally well in indoor locations when based on special markers. Whereas AR applications are initially developed for adding artificial figures into real world views, a camera with proper configurations can detect the precise positions and orientations of its special markers relative to the camera. This information is inversely used to calculate the location of the camera if the marker is stationary, so they are preferred in indoor navigation when we aim to track the viewer \citep{kim2008}. Although they function well, camera based devices are usually more expensive than BLE devices and they require the active movement of the camera to determine the location, which is not ideal if objects or people need to be passively tracked. Accordingly, La Delfa \etal~compare different marker technologies for environmental suitability \citep{ladelfa2016}. Xavier \etal~utilize markers for SLAM \citep{xavier2017}, whereas Ujkani \etal~perform sensor calibration with AR markers in a robotic cell achieving accuracies of 5-10 cm \citep{ujkani2018}.

In the wireless positioning community, AR positioning techniques are used hybridly to enhance the performance of the former systems as wireless signals are prone to yield high error rates \citep{alnabhan2014,koc2019}. Amongst these studies, Byrne \etal~propose a methodology for collecting RSSI information and accelerometer data using markers placed on the ground and cameras mounted on people to collect the ground truth \citep{byrne2018}.

We propose a framework that makes use of an AR-based localization system to collect precisely annotated RSSI data in both time and space. The system estimates positions of a specially designed camera-beacon bundle that captures video and RSSI data simultaneously, and is shown to implement a very accurate positioning with an error rate under 0.05 m and it can function alongside the BLE based localization without causing extra interference. With its low error rate, this method provides ground truth positions to evaluate or verify a pure wireless signal based localization system. After the postprocessing, we possess RSSI data points, each of which is annotated with its precise 3D position. Moreover, the proposed technique has a potential to greatly speed up the calibration procedure and decrease or fully automatize the workload of fingerprinting and radio frequency map generation.

As an end product, we also provide a publicly available dataset collected using the proposed method. The dataset includes position annotated RSSI data of several trajectories alongside the preprocessed probabilistic radio frequency map estimations in fingerprint or grid formats \footnote{The dataset is publicly available with DOI: 10.34740/KAGGLE/DS/1662453, and the framework is public in Github as ``Position-Annotated-BLE-RSSI-Dataset''.}.

In Section~\ref{sec:method}, the methods for data selection, filtering and synchronization that make up the position annotated wireless data collection framework are given. We describe our experimental setup and dataset specifications in Section~\ref{sec:exp}. The results supporting our proposed framework are reported in Section~\ref{sec:res} and we conclude with future insights in Section~\ref{sec:conc}.

\section{Data Annotation Methodology}
\label{sec:method}

\begin{figure}[!ht]
 \centering
\includegraphics[width=.9\linewidth]{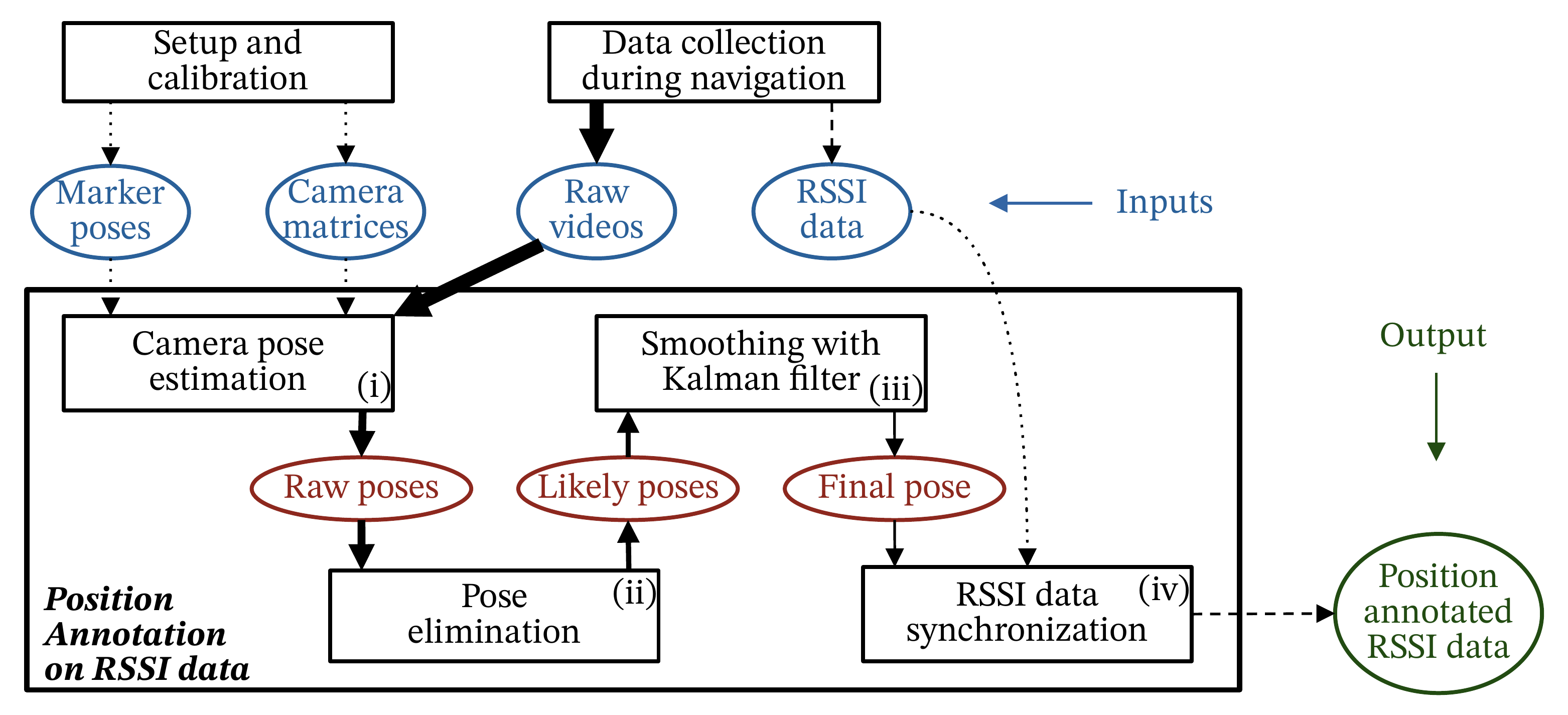}
\caption{Position annotation on RSSI data: we first (i) estimate the camera poses in the video frames using the marker poses and the camera calibration matrices, and then (ii) filter the likely poses by eliminating highly irrelevant estimations. The likely poses are (iii) smoothed using a properly designed Kalman filter to find a final pose for each frame, which is finally (iv) synchronized with the simultaneous RSSI data in time to obtain a position annotated RSSI dataset for each experiment in the area. The arrow weights denote the data load coarsely.\label{fig:pose_estimation}}

\end{figure}

Data annotation methodology comprises of four main consecutive sections as seen in \figurename~\ref{fig:pose_estimation}. With the camera calibration matrices, premeasured poses (positions and orientations) of the AR markers, timestamped RSSI readings and accompanying video frames in hand, we want to annotate each RSSI reading on a trajectory with a precise position estimated using the videos     and marker information.

Before going through the data annotation procedure, we first present the mathematical notations, definitions and necessary conversions in 3D space in Section~\ref{sec:notations}. In Section~\ref{sec:camera_poses} we employ the ArUco marker detector to estimate marker poses relative to the camera(s) \citep{romeroramirez2018,garridojurado2016}. With the defined mathematical foundations, estimated marker poses relative to the cameras are combined with the marker poses relative to the world to obtain the pose of the camera-beacon bundle in the world frame. In Section~\ref{sec:elimination}, we apply pruning techniques to eliminate the misleading estimations in a batch of pose estimations, which we observe to be very noisy due to erroneous marker detections. We apply a Kalman filter specially tailored for this problem to obtain the most relevant pose given a batch of poses in Section~\ref{sec:kalman}. Finally, in Section~\ref{sec:synchronization} these smoothed precise camera poses are synchronized in time with the RSSI readings.

\subsection{Notations, definitions and conversions}
\label{sec:notations}

Throughout this article, vectors are represented by bold lowercase letters (e.g. $\boldsymbol{\tau}$ or $\boldsymbol{\theta}$), while bold capital letters represent matrices (e.g. $\mathbf{R}$). The variables, \markeri, \camera~and \world~denote the entities, a marker, the camera-beacon bundle and the world, respectively. We represent the position (translation) and the rotation (orientation) vectors with \tvec{i}{jk} and \rvec{i}{j} respectively. The superscript on the left, $i$, is for the frame the vector is defined in, and the subscript on the right, $j$, is for the entity the vector belongs to. For the position vectors, a double subscript $jk$ shows the start and the end points. If the start position is the frame origin itself, we omit the first entity, $\tvec{j}{jk} \equiv \tvec{j}{k}$. For example, \rvec{\camera}{\markeri} defines the rotation vector of the marker \markeri, in the camera frame, \camera, and \tvec{\world}{\camera \markeri}, specifies a vector from the camera \camera~to the marker \markeri~in the world frame \world.

We use the rotation matrices to convert vectors from a predefined coordinate frame to another. A position vector defined in the frame $i$ is multiplied with the rotation matrix, \rmat{k}{i}, from the left to be converted into the coordinate frame $k$. Note that with a proper multiplication, the coordinate frames in the diagonals are eliminated, ($i$ in (\ref{eq:mult_vec})).
\begin{align}
\tvec{k}{j} = \rmat{k}{i}~\tvec{i}{j}
\label{eq:mult_vec}
\end{align}

Similarly, if we have a rotation matrix from the frame $i$ to $j$ and another one from $j$ to $k$, we can obtain transitively the rotation matrix from $i$ to $k$ by multiplying them as in (\ref{eq:mult_rot}). Note the elimination of the frame $j$.
\begin{align}
\rmat{k}{i} = \rmat{k}{j}~\rmat{j}{i}
\label{eq:mult_rot}
\end{align}

Since rotation matrices have a rank of three, a rotation vector is a convenient and compact representation of a rotation matrix. Conversion from a rotation vector to a rotation matrix is straightforward with the use of Rodrigues' formula (\ref{eq:rodrigues}), as follows:
\newcommand{\rvecind}[1]{\ensuremath{r_{#1}}}
\newcommand{\rvecunit}{\ensuremath{\boldsymbol{r}}}
\begin{align}
\begin{split}
\rmat{}{} & = Rodrigues(\rvec{}{})\\
& =  \identity \cos(\theta) + (1- \cos(\theta)) \rvecunit \rvecunit^T + \sin(\theta) \left [ \begin{matrix}
 0 & -\rvecind{z} & \rvecind{y}\\
 \rvecind{z} & 0 & -\rvecind{z}\\
 -\rvecind{y} & \rvecind{x} & 0
 \end{matrix}
\right ]
\end{split}
\label{eq:rodrigues}
\end{align}
where $\theta = \|\rvec{}{}\|$, is the norm of the rotation vector, $\rvecunit = \rvec{}{}/\theta$, the normalized rotation vector, and $\rvecind{i}$ are the individual orientation components for the axes in 3D. If we have a rotation vector of the entity $i$ defined in the coordinate frame $j$, we can obtain the associated rotation matrix from $i$ to $j$, using the Rodrigues' rotation formula \citep{liang2018}:
\begin{align}
\rmat{j}{i} &= Rodrigues(\rvec{j}{i})
\label{eq:Rodrigues}
\end{align}

The inverse of the Rodrigues' formula is also available, but with a more complicated setting:
\newcommand{\sRod}{s}
\newcommand{\cRod}{d}
\newcommand{\SRod}{S_{1/2}}
\newcommand{\vRod}{\mathbf{v}}
\newcommand{\ARod}{\mathbf{A}}
\newcommand{\rhoRod}{\boldsymbol{\rho}}
\begin{align}
\begin{split}
\rvec{}{} &= Rodrigues^{-1}(\rmat{}{})\\
& = \left \{
             \begin{matrix*}[l]
             \rvec{}{}  = 0, & \text{if }\sRod = 0\text{ and }\cRod = 1\\
             \rvec{}{} = \SRod \left (\frac{\vRod}{\|\vRod\|}\pi \right ), & \text{if }\sRod = 0\text{ and }\cRod = -1\\
             \rvec{}{} = \frac{\rhoRod}{\sRod} \theta, & \text{if }\sin \theta \neq\ 0\\
             \end{matrix*}
\right .
\end{split}
\end{align}
where $\ARod = \frac{\rmat{}{} - \rmat{}{}^T}{2}$, $\rhoRod = [a_{32}~a_{13}~a_{21}]^T$, $\sRod = \| \rhoRod \|$, $\cRod = (\text{tr}(\rmat{}{}-1))/2$, $\theta = \arctan_2(\sRod, \cRod)$, $\vRod$ is a nonzero column of $\rmat{}{} + \identity$ and $\SRod(\rvec{}{}) = -\rvec{}{}, \text{if }\|\rvec{}{}\| = \pi \text{ or }\SRod(\rvec{}{}) = \rvec{}{},\text{ otherwise}$. Both of these formulas can be easily employed through the OpenCV library. If we have a rotation matrix from the coordinate frame $i$ to the frame $j$, we can obtain the associated rotation vector of the entity $i$ relative to the coordinate frame $j$, using the inverse of the Rodrigues' rotation formula:
\begin{align}
\rvec{j}{i} &= Rodrigues^{-1}(\rmat{j}{i})
\end{align}

We denote the positions of stationary markers as \tvec{\world}{\markeri}, and their orientations as \rvec{\world}{\markeri}. When markers are detected and recognized in a video frame, the pose estimator supplied by the ArUco suite estimates the position vectors pointing to the center of the marker, \tvec{\camera}{\markeri}, and rotation vector, \rvec{\camera}{\markeri}, relative to the camera coordinate frame.

\subsection{Camera pose estimation}

\label{sec:camera_poses}

\begin{figure}[!ht]
 \centering
\includegraphics[width=.7\linewidth]{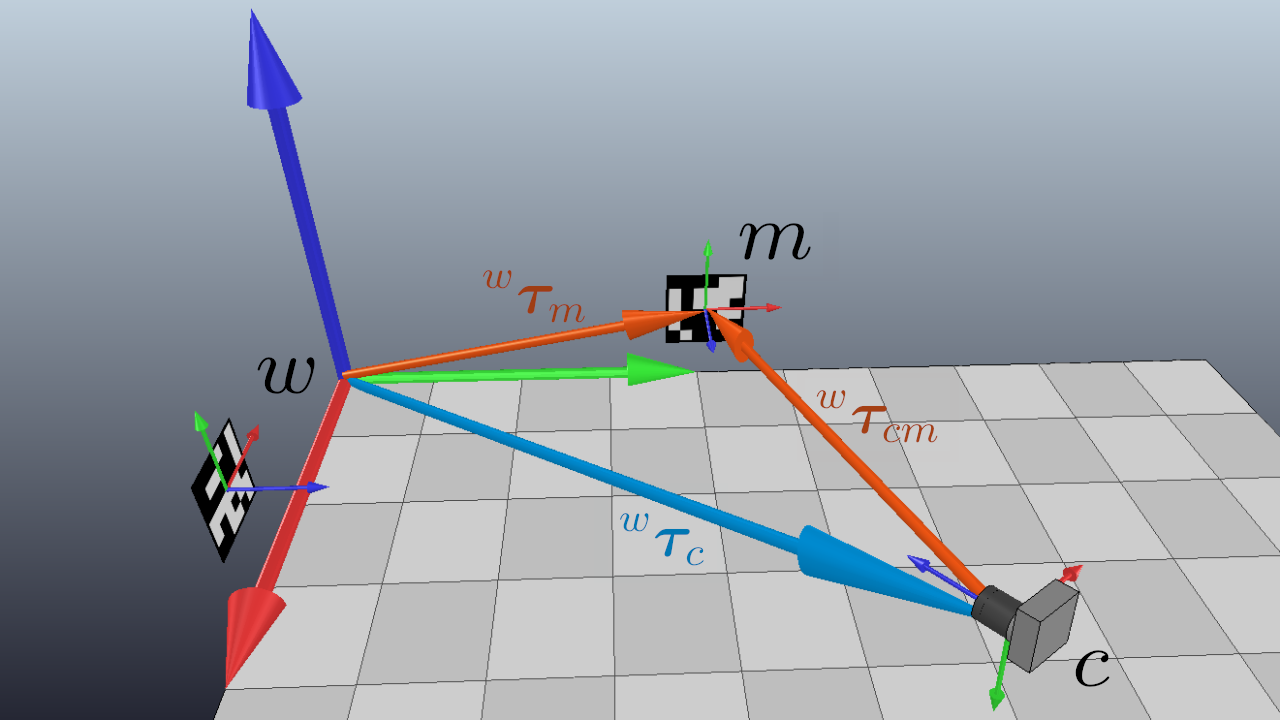}
\caption{The position vectors and the coordinate frames: \world, \camera~and \markeri; for the world, the camera and the marker frames respectively. The axes, $x$, $y$ and $z$ are coded with red, green and blue arrows respectively. We are to compute the cyan vector, given the orange vectors.}
\label{fig:frames}
\end{figure}

We are interested in extracting the poses of the camera-beacon bundle, which we will denote (\tvec{\world}{\camera}, \rvec{\world}{\camera}). The ArUco suite provides the marker detector routines. With properly calibrated cameras, we can estimate the poses of the markers relative to the camera coordinate system, denoted as (\tvec{\camera}{\markeri}, \rvec{\camera}{\markeri}). We relate the latter information to the former using the premeasured marker poses relative to the world frame, (\tvec{\world}{\markeri}, \rvec{\world}{\markeri}) (see Section~\ref{sec:markers} for details on markers). For the positions part, we do:
\begin{align}
\begin{split}
\tvec{\world}{\camera} & = \tvec{\world}{\markeri} - \tvec{\world}{\camera \markeri}\\
& = \tvec{\world}{\markeri} - \rmat{\world}{\markeri}~ \rmat{\markeri}{\camera}~\tvec{\camera}{\camera \markeri}\\
& = \tvec{\world}{\markeri} - \rmat{\world}{\markeri}~\left (\rmat{\camera}{\markeri} \right )^T~\tvec{\camera}{\markeri}
\end{split}
\label{eq:position}
\end{align}
where $\rmat{\world}{\markeri} = Rodrigues(\rvec{\world}{\markeri})$ and $\rmat{\camera}{\markeri} =  Rodrigues(\rvec{\camera}{\markeri})$. The computation consists of a mere redefinition of the position vector from the camera to the marker in the world coordinate frame, \tvec{\world}{\camera \markeri}, and its subtraction from the marker position in the same coordinate frame, \tvec{\world}{\markeri} as seen \figurename~\ref{fig:frames}.

We can also easily obtain the rotation vector using the rotation matrix from the world frame to the camera frame, \rmat{\world}{\camera}. Whereas this matrix can be used to convert a vector from camera coordinate system to the world coordinate system, it also defines how the camera is oriented within the world frame. The rotation matrix was already computed as an intermediate step of the position computation (\ref{eq:position}).  Employing the inverse of the Rodrigues' formula, we convert the rotation matrix into the rotation vector of the camera in the world frame (\ref{eq:orient}) as follows:
\begin{align}
\begin{split}
\rvec{\world}{\camera} &= Rodrigues^{-1}(\rmat{\world}{\camera})\\
& = Rodrigues^{-1}\left (\rmat{\world}{\markeri}~\left (\rmat{\camera}{\markeri} \right )^T \right )
\end{split}
\label{eq:orient}
\end{align}

The conversions and computations above are repeated for each detected marker in an image frame. Since we construct our detection setup aiming to capture several markers at a time, we expect to estimate multiple camera pose vectors at each frame, (\tvec{\world}{\camera}, \rvec{\world}{\camera}), namely the raw poses.

\subsection{Pose elimination}
\label{sec:elimination}

\begin{figure}[!ht]
\centering
\includegraphics[width=.44\linewidth]{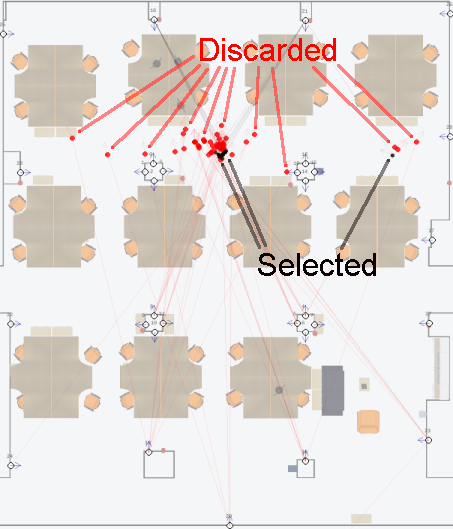}\hspace{10pt}
\includegraphics[width=.44\linewidth]{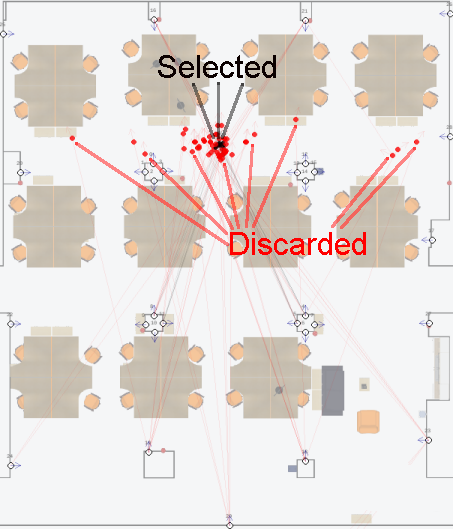}
\caption{Position elimination strategies: close marker selection (left) and outlier elimination (right). Black points are selected for the next stage, while the red points are eliminated. \label{fig:elimination}}
\end{figure}

The variances of the camera pose estimations are observed to be large due to noisy marker pose detections caused by minor errors in manual measurements, faulty camera matrices and coefficients, motion blur or varying lighting conditions. The combination of these factors lead to inaccurate rotation matrices, which in turn accumulate to varying positions for the camera. Before  directly applying a filtering algorithm on the raw poses, we employ two elimination strategies (see \figurename{~\ref{fig:elimination}}).

At each video frame, we expect to recognize multiple markers. We use two cameras to expand the field of view, so if one camera misses the markers, the other one captures possibly more than one marker. Moreover, to make sure that there exist multiple marker readings, we process \frames~video frames in one shot. The traversed distance of a batch, $\distance_\frames$, for the camera-beacon bundle moving at a constant speed of $v$ can be computed with $\distance_\frames = v \frames / (2 fps)$, where $2fps$ is the frame capture speed of a two camera system. For example, with $\frames = 15$, $fps = 60$, and navigating at $v = 0.35~m/s$, each batch of frames corresponds to about $\distance_\frames = 0.044$ meters.

We observe that camera poses from distant markers can be more misleading. This defect is due to the number of pixels of the detected marker image. As the distance increases, the lack of pixels lead to larger measurement errors. Therefore, as long as multiple poses are estimated in a time window of \frames~frames, the poses based on distant markers may be discarded. We set a closeness parameter, \selection, that takes into account the closest \selection~marker readings. This is a quick technique that sorts the detected markers in a frame with respect to their distances to the camera and keeps only the least \selection~distant poses.

Another elimination is achieved by an outlier detection test. In a set of pose estimations, we eliminate the outliers with respect to the distances to a consensus point \citep{fischler1981}. The parameter \outlier, is defined to determine the number of remaining poses after consecutive outlier detection iterations. This technique is merely a slow backward subset selection method that is repeated in a greedy manner.

Employing the elimination methods with proper values for the parameters, \selection~and/or \outlier, we prune the raw poses into a smaller set of poses, namely the likely poses.

\subsection{Kalman filter for smoothing}
\label{sec:kalman}

Even though some marker detections are eliminated, we still receive erroneous estimations. Error fluctuations can be controlled by employing appropriate filtering mechanisms. The current section is linked to the previous section through pose vector estimates of the camera-beacon bundle, which will be our observations in the filtering design: $\obs_t = (\tvec{\world}{\camera}, \rvec{\world}{\camera})_{t}$. We model this new problem as a state space model, in which the observed poses, $\obs_t$, are assumed to be generated noisily from the latent poses, $\point_t$. The latent pose vector of the tracked object, $\point_t$ at time $t$, is a function of the pose vector at a previous time, $\point_{t-1}$. This function is also called the motion model that stands for the dynamics of the object.

The transition densities are assumed not to follow explicit motion models. Instead, we assume that $\point_t$~resides close to $\point_{t-1}$. Observation densities are designed similarly. $\obs_t$ should not be far away from $\point_t$. With a prior distribution at the initial time, $p(\point_0)$, we are to find the filtering distribution $p(\point_t|\obs_{1:t})$. These assumptions can be realized by forming transition and measurement densities with Gaussians (\ref{eq:hmm}), whose noise covariance matrices will be denoted as $\transN_t$ and $\obsN_t$ respectively.
\begin{align}
\begin{split}
\point_0 & \sim p(\point_0)\\
\point_t & \sim \mathcal{N}(\point_{t}; \point_{t-1}, \transN_t)\\
\obs_t & \sim \mathcal{N}(\obs_t; \point_{t}, \obsN_t)
\end{split}
\label{eq:hmm}
\end{align}

The dimensions of the latent space and the measurement space are assumed to be independent, so we use diagonal matrices. The values of the covariance matrices are chosen to compensate for unexpected high errors in measurements \citep{wang2015,chen2016}. The exact inference for the filtering distribution becomes tractable, and can be solved with the standard Kalman filter:
\begin{align*}
\text{Predict:} & \\
\hat{\point}_{t} & = \transM \point_{t-1}\\
\hat\predict_{t} & = \transM \predict_{t-1} \transM^T + \transN_t\\
\text{Update:} & \\
\kalmangain_t & = \hat\predict_{t}  \obsM^T  (\obsM \hat\predict_{t} \obsM^T + \obsN_t)^{-1}\\
\point_{t} & = \hat\point_{t} + \kalmangain_t (\obs_t - \obsM \hat\point_{t})\\
\predict_{t} &= (\identity_d - \kalmangain_t \obsM) \hat\predict_{t}
\end{align*}
where transition and observation matrices, $\transM$ and $\obsM$, are chosen as identity matrices ($\identity_d$), $\predict_{t}$ are the estimated covariance matrices, and $\kalmangain_t$ are the optimal Kalman gains. We obtain $\point_{t}$ as the pose estimations, namely the final poses.

\subsection{RSSI data synchronization}
\label{sec:synchronization}

After the filtering operation, timestamped pose data are obtained, but there still exist problems in synchronizing the timestamped RSSI data to the timestamped position data.

\begin{figure}[!ht]
  \centering
  \subcaptionbox{Detecting covering visually.\label{fig:sync_hands}}{
  \includegraphics[width=.396\linewidth]{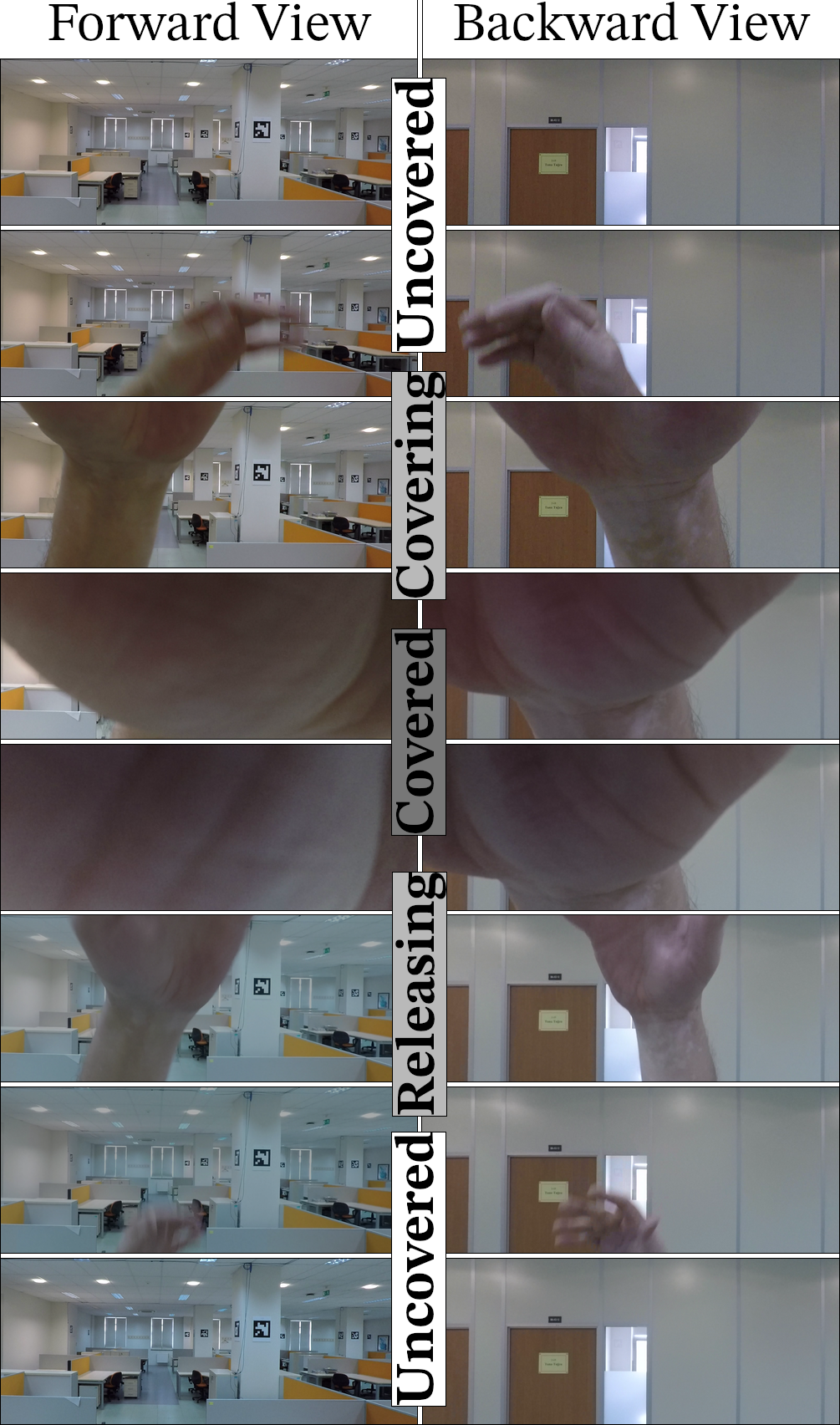}}
  \subcaptionbox{Detecting covering in RSSI readings.\label{fig:sync_signal}}{
  \includegraphics[width=.468\linewidth]{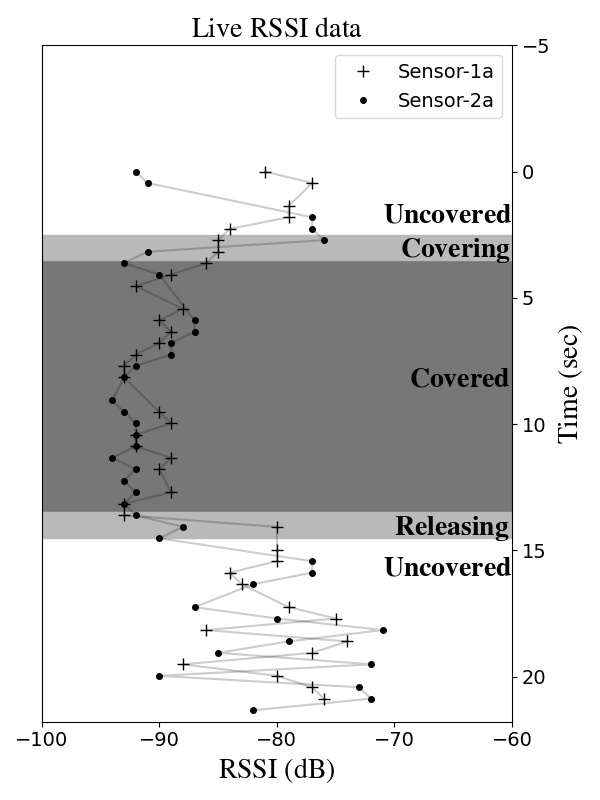}}
  \caption{Synchronization of video frames with RSSI data in time. We apply artificial disturbances on the videos and the RSSI data by our hands. We observe three stages: the bundle is uncovered (white); the hands are hold on the bundle (gray) and the transition stage where the hands are moving to cover or release the bundle (light gray).
  \label{fig:sync}}
\end{figure}

The first problem is that the cameras cannot be synchronized to a precise universal clock. We overcome this problem by adding artificial disturbances that affect both the camera views and the RSSI data so that a synchronization may be possible. After various unsuccessful Faraday cage-like attempts, we found that a pair of hands would be practical to give disturbances to electromagnetic signals and visual data. We cover and release the camera-beacon bundle with two hands at the beginning of each experiment. This hand movement can be clearly seen in the video frames as shown in \figurename{~\ref{fig:sync_hands}}. Because body parts absorb electromagnetic signals, we observe that this method creates a significant reduction in the RSSI data. The artificial disturbance can easily be detected in RSSI readings as seen in \figurename{~\ref{fig:sync_signal}}. In the figures, we divide this hand movement into four color-coded stages. We first cover the camera-beacon bundle quickly (light gray). When the bundle is covered with hands, we are in the gray region where the cameras' views are largely blocked and we observe that strong RSSI readings coarsely between -85 to -65 decibels are lost because of the absorption due to the hands. Then we release (light gray) the bundle as quickly as possible to reach the white region where the camera views are no longer blocked and the RSSI readings return to their normal course. We select the timestamp of the second light gray region that points to the change point between the white region and the gray region. We modify the data streams to start from this selected timestamp, and we are left with synchronized video frames and RSSI data.

\begin{figure}[!ht]
    \centering
    \includegraphics[width=.44\linewidth]{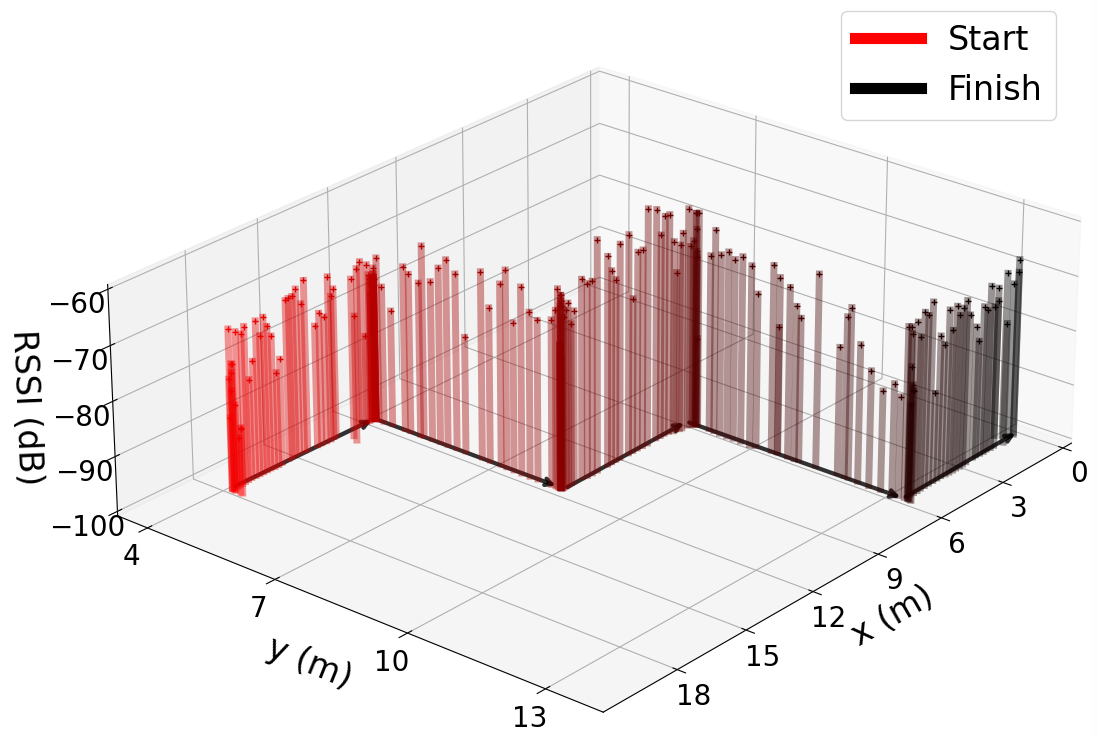}
    \includegraphics[width=.44\linewidth]{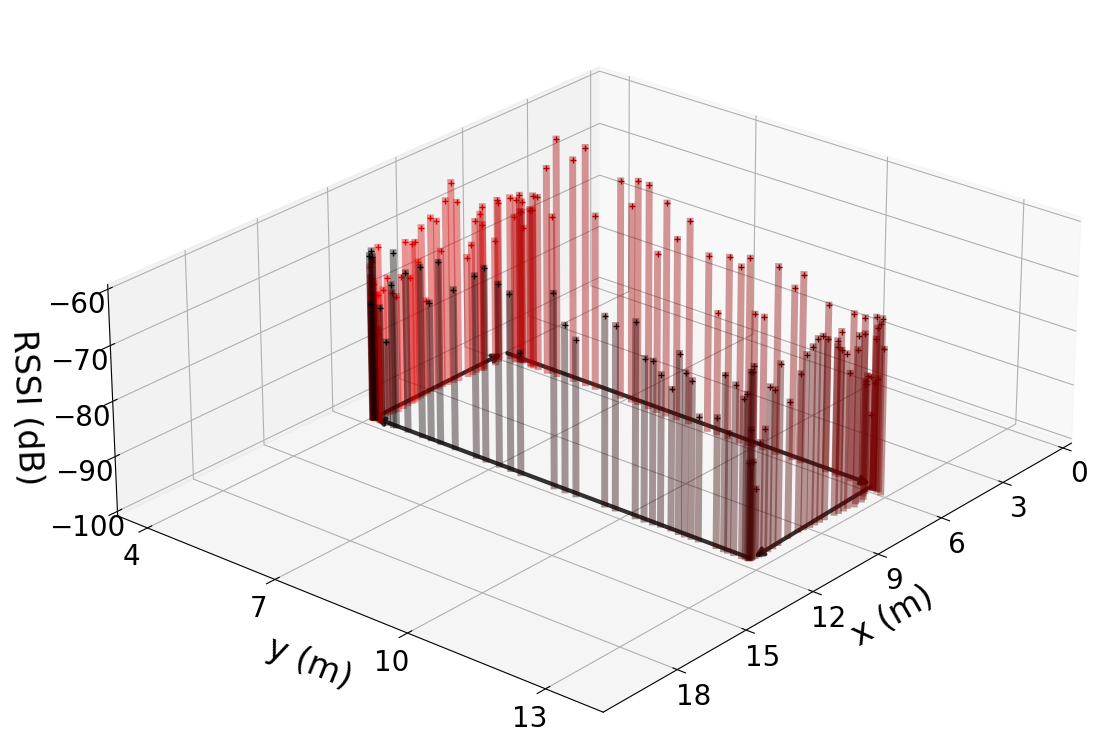}
    \caption{RSSI data overlaid on the position data: time flows from red bars to black bars.}
    \label{fig:rssi_overlaid}
\end{figure}

Whereas the data streams are set to begin at the same time, RSSI timestamps and video timestamps do not match exactly. Secondly and obviously, we have to unify the two timestamps. RSSI data is far less frequent than the pose data (2 Hz vs 10-30 Hz). To coincide the two data types in time, we perform a linear interpolation of the video frame timestamps at the RSSI timestamps, and we label each RSSI timestamp with the linearly interpolated pose data as its ground truth. After the time synchronization stage, we have a dataset of RSSI data points annotated with the position it is captured on. Two of the trajectories from our dataset are displayed in \figurename{~\ref{fig:rssi_overlaid}}.

\section{Data Collection Setup}
\label{sec:exp}

\begin{figure*}[!ht]
\centering
\includegraphics[width=.48\linewidth]{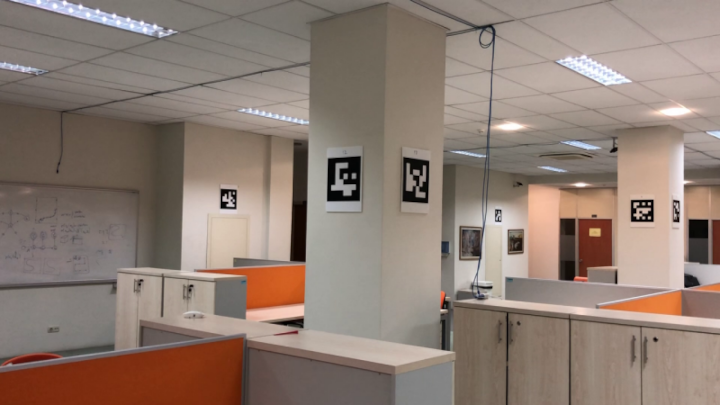}
\includegraphics[width=.48\linewidth]{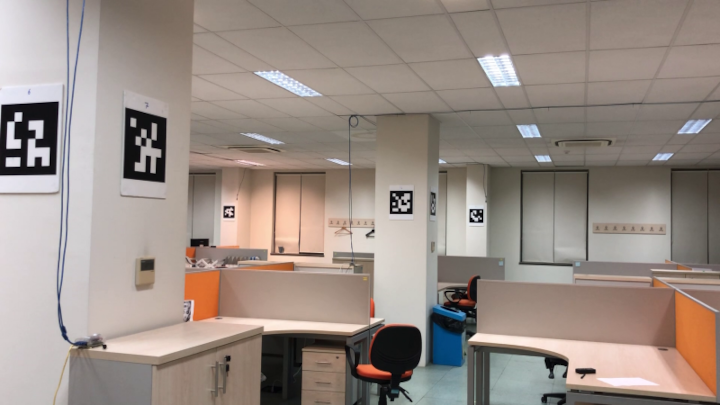}
\caption{Sample images from the area with markers, extension cables, BLE adapters, computers and other obstacles. \label{fig:area}}
\end{figure*}

The aim of the data collection setup is to exploit as many sensors as possible and build a BLE sensor network that covers the whole area at a minimal cost, and finally build a dataset with the highest information within. We use a set of BLE beacons and adapters, low profile computers, motion cameras and visual markers installed in an indoor area of size $20.66\times17.64$ m${}^2$. Intended originally as an office space for fellow graduate students, the area also includes columns, walls and furniture that cause multipath interference on the wireless signals (see \figurename{~\ref{fig:area}}).

\begin{figure}[!ht]
\centering
\subcaptionbox{Illustration of the sensor distribution: yellow rectangles are the computers, colored circles the sensors and the gray lines are the USB cables that link the sensors to the computers. \label{fig:sensor_distribution} }{\includegraphics[width=0.358\linewidth]{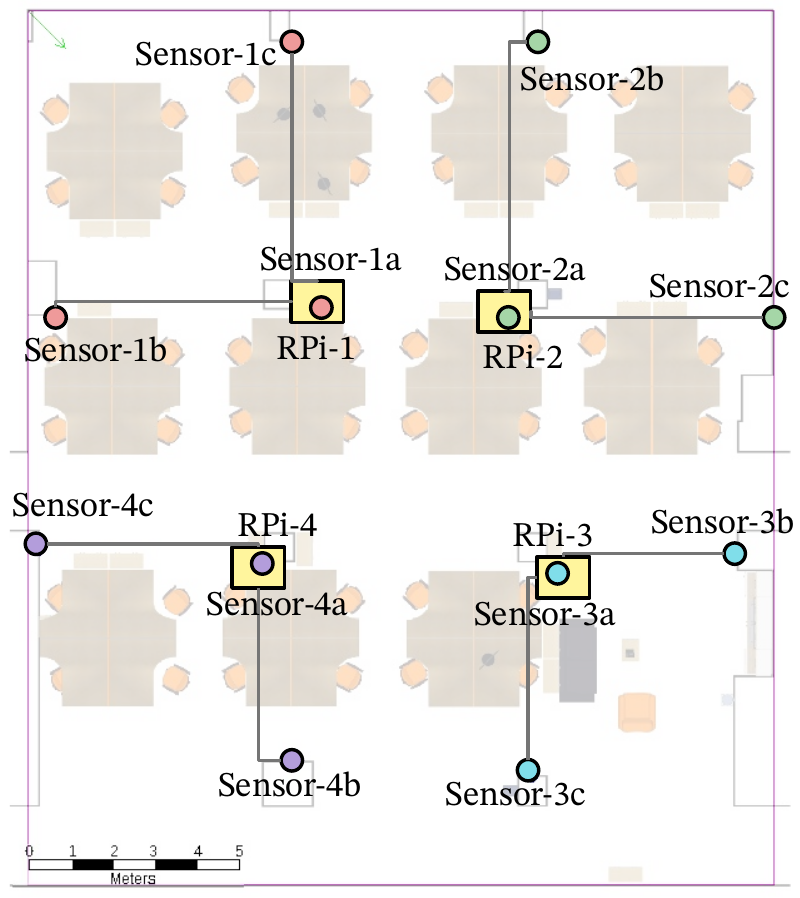}}\hspace{5px}
\subcaptionbox{The navigable platform (left); the beacon-camera bundle of a POI beacon and two GoPro Hero 4 motion cameras (top middle); a Raspberry Pi 3 Model B+ computer (down middle); Logilink Bluetooth adapter (top right); and a KLPRO KLLMZ60 laser distance meter (down right). \label{fig:cam_beacon}}{
  \begin{minipage}{0.195\linewidth}
      \includegraphics[width=\linewidth]{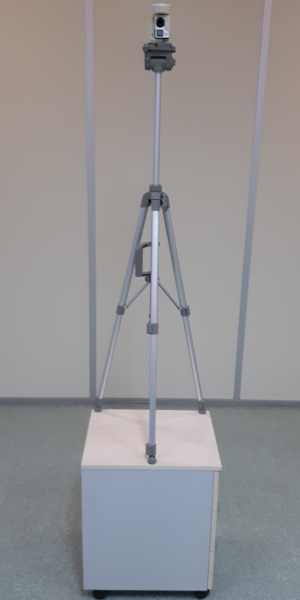}
  \end{minipage}\hspace{1px}
  \begin{minipage}{0.390\linewidth}
      \includegraphics[width=\linewidth]{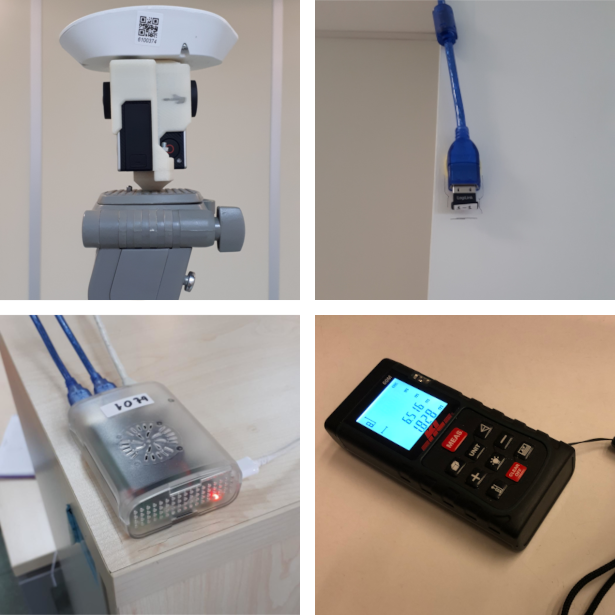}
  \end{minipage}}
\caption{Hardware setup}
\end{figure}

\subsection{Hardware setup}

We consider that an efficient configuration of the sensor positions in terms of data quality can be achieved by putting them on the walls surrounding the test area. However, there appear difficulties in both powering the sensors and reading data from them with more than 10 meters of the extension cables. Because of the cable length constraints, in our design, each two sensors are controlled by one computer. To cover the whole area as equidistantly as possible, we install 8 Logilink Bluetooth 4.0 adapters on the walls, and use the 4 onboard Bluetooth embedded chips of the computers, making 12 Bluetooth sensors (Sensor-$ij$) in total as seen in \figurename{~\ref{fig:sensor_distribution}}. The computers are placed on the cabinets to ensure that their embedded sensors have a standard elevation of one meter (see \figurename{~\ref{fig:cam_beacon}}). The USB cables are attached to the ceiling to minimize possible signal interferences. The adapters are attached on the walls elevated about two meters from the floor. In this configuration the extension cables are tested to supply enough power to the adapters. The computers responsible for the data capture system are four Raspberry Pi 3 Model B+ single board computers (RPi-$i$), which are based on Broadcom BCM2837B0, Cortex-A53 (ARMv8) 64-bit quad-core processors. They run Raspbian Stretch operating system, on which the Robot Operating System (ROS) Melodic middleware is installed \citep{quigley2009}. The computers scan the BLE packets emitted by the transmitters and publish them into the local area network (see \figurename{~\ref{fig:cam_beacon}}).

We use one sample of POI Beacons as the BLE signal transmitter, which is constructed upon Nordic Semiconductor multi-protocol SoC and 32-bit Arm Cortex M0 processor with 256 kB flash and 16 kB on chip. POI Beacons use the Bluetooth 4.0 BLE protocol and iBeacon protocol, and have a signal range of 70 meters. The beacons send two BLE identity packets per second and have a running time up to 4 years with two standard AA batteries (see \figurename{~\ref{fig:cam_beacon}}).

A 3D printed custom case, which we call ``the camera-beacon bundle'', holds the BLE beacon and two GoPro Hero 4 motion cameras together (see \figurename{~\ref{fig:cam_beacon}}). The cameras are capable of capturing images with a resolution of $1920 \times 1080$ at 60 fps. The special design of the custom case lets the cameras stay in such a formation that they are placed back to back, and their lenses are aligned facing opposite directions. One camera stays upside down with respect to the other one. The design also minimizes the translations between camera coordinate frames and the beacon, provides a large view angle in order to detect as many markers as possible at an instant, and maximizes the functionality of the cameras by allowing reaching both of the buttons. The cameras and the BLE beacon are deliberately put in the same place as we are to collect BLE data originating from the beacon and simultaneous videos that will be used to implicitly estimate precise positions of the mobile beacon. By this setup, we build a data collection infrastructure and then by processing the videos captured by the cameras, we estimate the positions of the camera-beacon bundle, so that the RSSI readings by the BLE beacon can be labeled with those positions. The labels will serve as ground truth for the future pure BLE based localization purposes.

The bundle is attached onto a tripod which in turn is mounted on a wheeled cabinet that allows smooth navigation in the test area (see \figurename{~\ref{fig:cam_beacon}}). The beacon height is set to be 2 meters to obtain a clear view of the test area in terms of both visual data and radio frequency data. The tripod functionality also allows us to switch to different alignments of the camera-beacon bundle, from which we prefer horizontal alignment to maximize the view angle, hence, the number of detectable markers, and to minimize the interference on the beacon due to the cameras.

The distances are rigorously measured with a KLPRO KLLMZ60 laser distance meter with a working range of 60 m and an accuracy of $\pm0.002$ meters (see \figurename{~\ref{fig:cam_beacon}}).

\begin{figure*}[!ht]
\centering
\subcaptionbox{ArUco marker samples (top) and the calibration chart (bottom).\label{fig:markers}}{
\begin{minipage}{0.415\linewidth}
\vspace*{\fill}
\includegraphics[width=\linewidth]{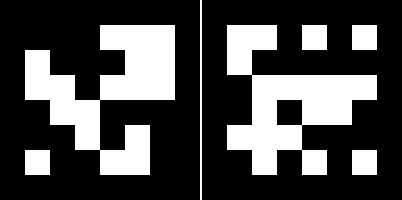}
\par\vspace{.1em}
\includegraphics[angle=90,width=\linewidth]{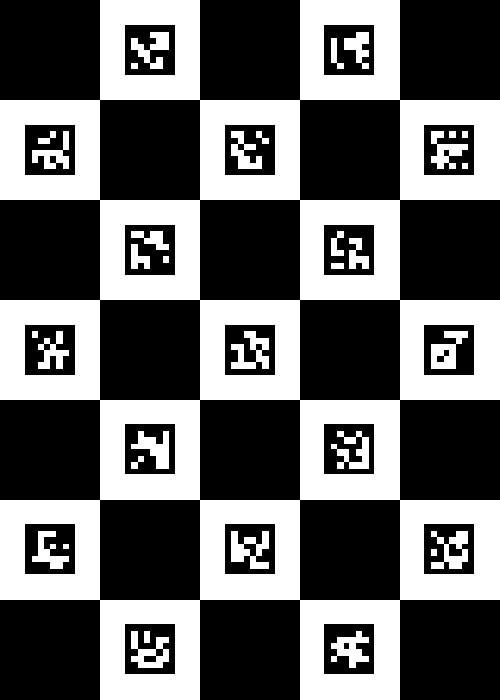}
\end{minipage}}
\hspace{10pt}
\subcaptionbox{The markers in the test area as black dots and arrows.\label{fig:marker_plan}}{\includegraphics[width=.463\linewidth]{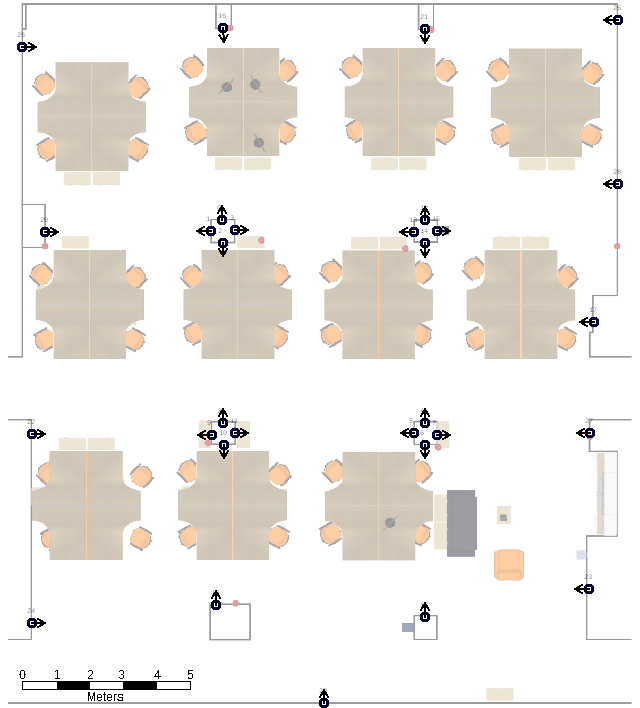}}
\caption{ArUco markers and their distribution in the test area.}
\end{figure*}

\subsection{Markers and camera calibration}
\label{sec:markers}

We choose 30 square markers from the dictionary of $6\times6$ matrix markers of the ArUco marker suite (\figurename{~\ref{fig:markers}}), and print them on hard paper with edges of 0.3 meters. The markers are attached onto the walls or columns where they can be viewed from almost any angle in the area (\figurename~\ref{fig:marker_plan}). We record the label and the pose (of the center) of each marker with respect to a predetermined world coordinate frame (\tvec{\world}{\markeri}, \rvec{\world}{\markeri}). The measurement of the position vector, \tvec{\world}{\markeri}, is straightforward, that is, we manually measure the distances to the walls of reference and to the floor with the help of the laser distance meter, but the computation of a rotation vector, \rvec{\world}{\markeri}, may be challenging. To ease this process, we install the markers on the walls or columns in such a way that rotations between the marker coordinates and the world coordinates can be easily computed by multiples of $\pi/2$. Specifically the $y$-axis of the markers are always aligned with the $z$-axis of the world coordinate system. To compute a rotation vector of a marker in world coordinates, we first compute the required rotation around the $x$-axis, and then around its $z$-axis. For example, for the marker \markeri~in \figurename{~\ref{fig:frames}}, we first rotate it around its $x$-axis by $\pi/2$, and then around its $z$-axis by $\pi/2$ again to coincide the marker coordinate frame with the world coordinate frame. The rotation vector of the marker \markeri~in the world coordinate system is then recorded as $\rvec{\world}{\markeri} = [\pi/2, 0, \pi/2]$.

The ArUco suite provides calibration procedures as well as generating the marker images and detecting them in still images. Using several views of the Charuco calibration chart \citep{an2018} (\figurename{~\ref{fig:markers}}), the camera matrix and the corresponding distortion coefficients are tuned. A highly accurate camera matrix is mandatory for precise pose detection of the markers.

The current positioning of two cameras allows us to have an almost spherical world view with the fish eye mode of the cameras, but we prefer the planar view mode of cameras to the fish eye mode for three reasons: (i) the calibration procedure is challenging for the fish eye view, (ii) the fish eye distortions have to be corrected before running the AR Marker detection algorithms as they have to be fed with planar markers, and (iii) we do not need to see ceilings and with the planar view mode, the cameras already have a large enough angle of view.

\subsection{Handling setup errors}

The errors in marker based localization are due to multiple factors that can be divided into two sets: human errors and environmental errors. Human errors include mostly errors in measuring the marker poses, reference points and unstable navigation. We can also put the camera calibration based errors into this set, which usually lack the necessary views for the calibration process. Environmental errors are due mostly to ambient light fluctuations.

The markers are installed with the following factors taken into consideration in order to minimize the errors due to human factors: (i) From any position with any orientation in the area, at least one marker should be detectable. Markers should be above the objects that can cause visual occlusion. (ii) Markers are attached in the places illuminated uniformly and sufficiently. (iii) Markers are placed with the help of a jig to ensure that they are at a fixed height and orientation to facilitate the height and rotation measurements.

We minimize the errors due to camera calibration by using several views of the Charuco calibration chart \citep{an2018} (\figurename{~\ref{fig:markers}}), by which the camera matrix and the corresponding distortion coefficients are tuned properly. To evaluate the calibration and detection performance before running the main set of experiments, we collect a preliminary set of still images. For this, the navigable platform is put in the area in front of each marker at different angles and distances, but the office has a raised floor that is prone to flexing while people are stationary or mobile in the proximity of the image capture device. The tilting or oscillation of the image capture device can have a small effect on the captured position information. For calibration and verification purposes, the authors waited until the capture platform was stabilised and image capture was triggered wirelessly to obtain the best results. Sample images can be seen in \figurename{~\ref{fig:area}}. By adjusting the camera matrix and the parameters to their best values, we achieve an empirical minimum mean squared distance error of 0.05 meters for all markers installed in the area. As the recent studies on BLE based localization techniques report localization errors no less than 1 meters for practical setups \cite{tomazic2021, obeidat2021}, with our controlled error rate at 0.05 meters, marker based localization is highly better than any other BLE or WiFi based localization method. Thus this technique can be used as the ground truth to evaluate the RF RSSI based localization performance.

The installation of the AR markers is a tedious and time consuming work, but this procedure can be eased by considering the following points: (i) Instead of sticking the markers on the walls, fixing them on portable stands at predetermined elevations will reduce the installation time significantly. Polycarbonate stands should be preferred to minimize wireless signal attenuation and interference due to these artificially introduced materials. The portable marker stands can be removed from the area quickly after the data collection stage. (ii) Since markers can be recognized from any angle, choosing common orientations like the multiples of $\frac{\pi}{2}$ will help greatly. With standard elevations and easy orientation labeling, the experimenter is left with measuring the marker positions in 2D coordinates. (iii) We use two motion cameras not to miss any marker at any time interval, but this can be regarded as overkill. A single camera will also be sufficient to detect the markers, but the focus of the camera should be static, as variable focal length will disrupt the marker pose estimations. (iv) Instead of mobile beacons, one can use a mobile sensor (i.e. computer or smartphone) to capture the BLE packets emitted by the beacons that furnish the area. Several other sensors can be introduced into the setup, leading to more data that can improve the position estimation accuracy.

\subsection{Data collection software setup}

For sharing data among multiple computers connected through Ethernet, we use the ROS middleware. ROS middleware is a collection of open-source frameworks that are intended to be used in robot software development. Besides the pure robotics services like low-level device control, hardware abstraction, package management and common robotics functionality; the middleware provides a message-passing service between processes running at different devices. The messages of sensor data, control, state, planning, actuator information are shared among processes, which are called nodes. The nodes can publish or subscribe to the topics where the messages are streamed. The middleware is developed to be platform and language independent.

\begin{figure}[!ht]
\centering
\includegraphics[width=\linewidth]{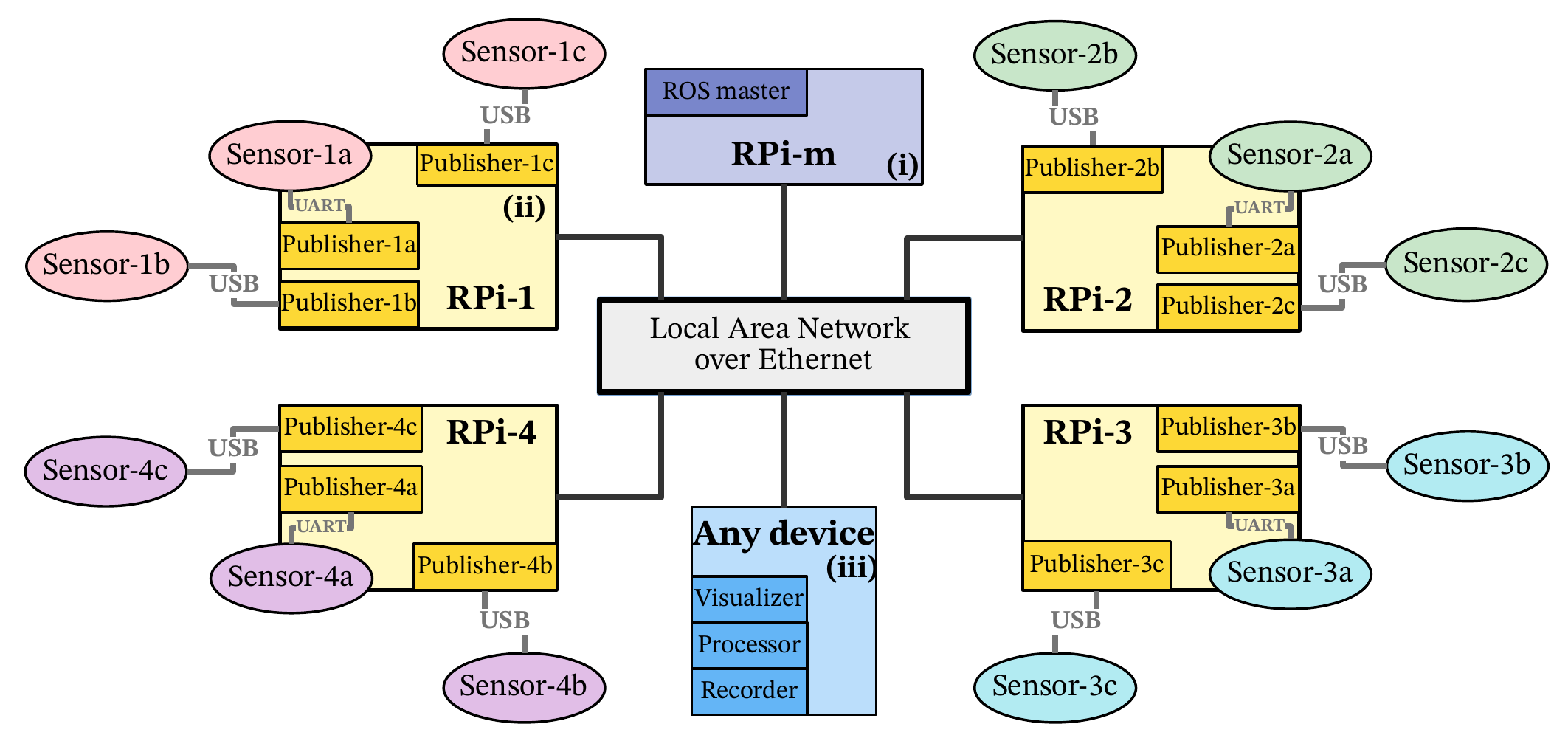}
\caption{Data collection and communication architecture: A ROS master computer (i) runs a master node that orchestrates and takes the role of a system access point for the other nodes. Dedicated RPi devices (ii) run a publisher node for each sensor to publish the sensor data into the system. Any other device (iii) can be used for live data visualization, collection and higher level processing.
\label{fig:ros_architecture}}
\end{figure}

\begin{figure}[!ht]
\centering
\includegraphics[width=.275\linewidth]{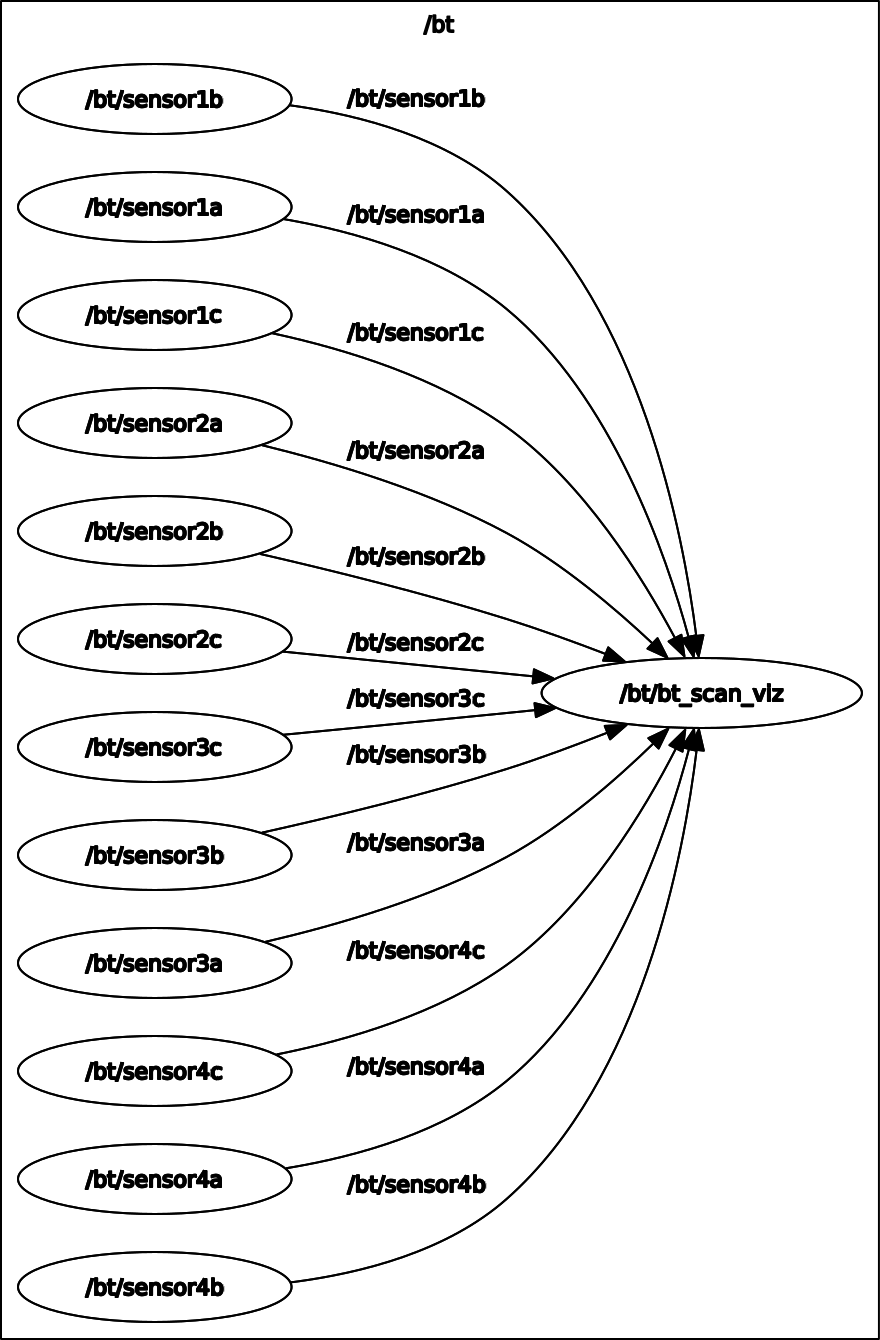}
\hspace{10pt}
\includegraphics[width=.488\linewidth]{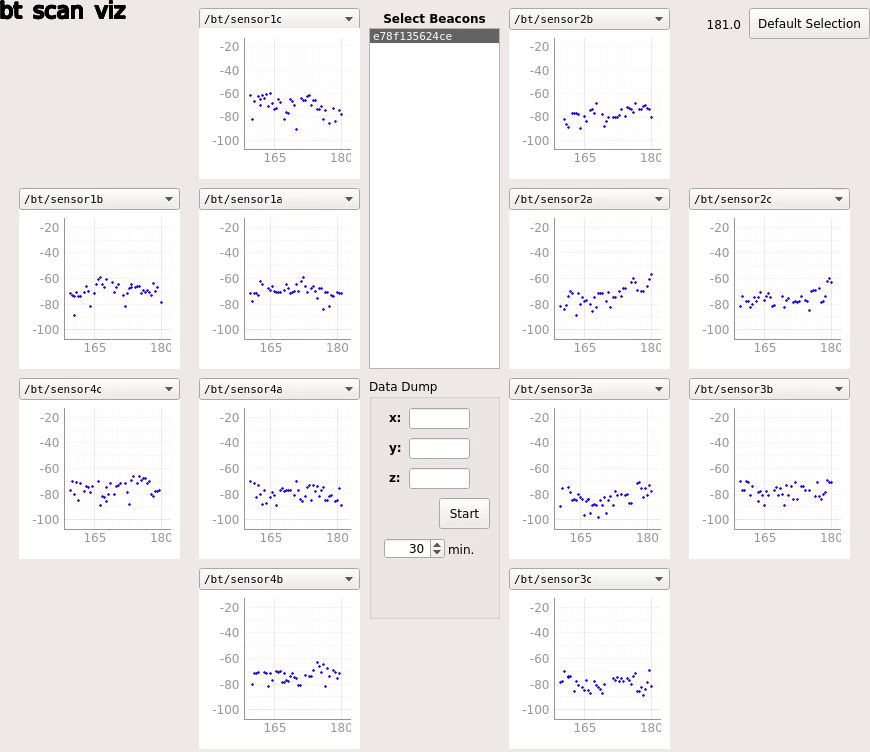}
\caption{RSSI data flow on ROS (left) and the live data visualizer (right). \label{fig:data_visualizer}}\end{figure}

In this work, the middleware is used for message-passing, visualizing and recording data. However, with its high and easy portability degree, the infrastructure allows plugging any node in real-time. The proposed architecture is open to improvement in the domain of wireless sensor networks. Our ROS architecture, as seen in \figurename{~\ref{fig:ros_architecture}}, has three main node types: we have a master node (i) for orchestration and an entry point for other nodes. We prefer to dedicate another computer, RPi-m, that runs the master node. This node is responsible for triggering the publishers in the network. The network can easily be expanded with other nodes. A node that wants to enter into the ROS network has to present itself to this master node. Each computer, RPi-$i$, runs three publisher nodes, Publisher-$ij$, (ii), which are responsible for scanning the corresponding Bluetooth devices, Sensor-$ij$, generate RSSI parameters from incoming Bluetooth packets, and publish these parameters as ros messages. The published messages include a timestamp, MAC address of the source beacon, MAC address of the sensor that captures the Bluetooth packet and the RSSI value in decibels. Finally, we have a visualizer (iii) that subscribes to the RSSI topic messages and displays them in live windows with respect to time. The visualizer is also responsible for recording the data in log files for each sensor. The master, publisher and visualizer nodes are running on Python-3 scripts. The visualizers also have a graphical user interface based on PyQt5 wrappers (see \figurename{~\ref{fig:data_visualizer}}).

\begin{figure}[!ht]
\centering
\includegraphics[width=.8\linewidth]{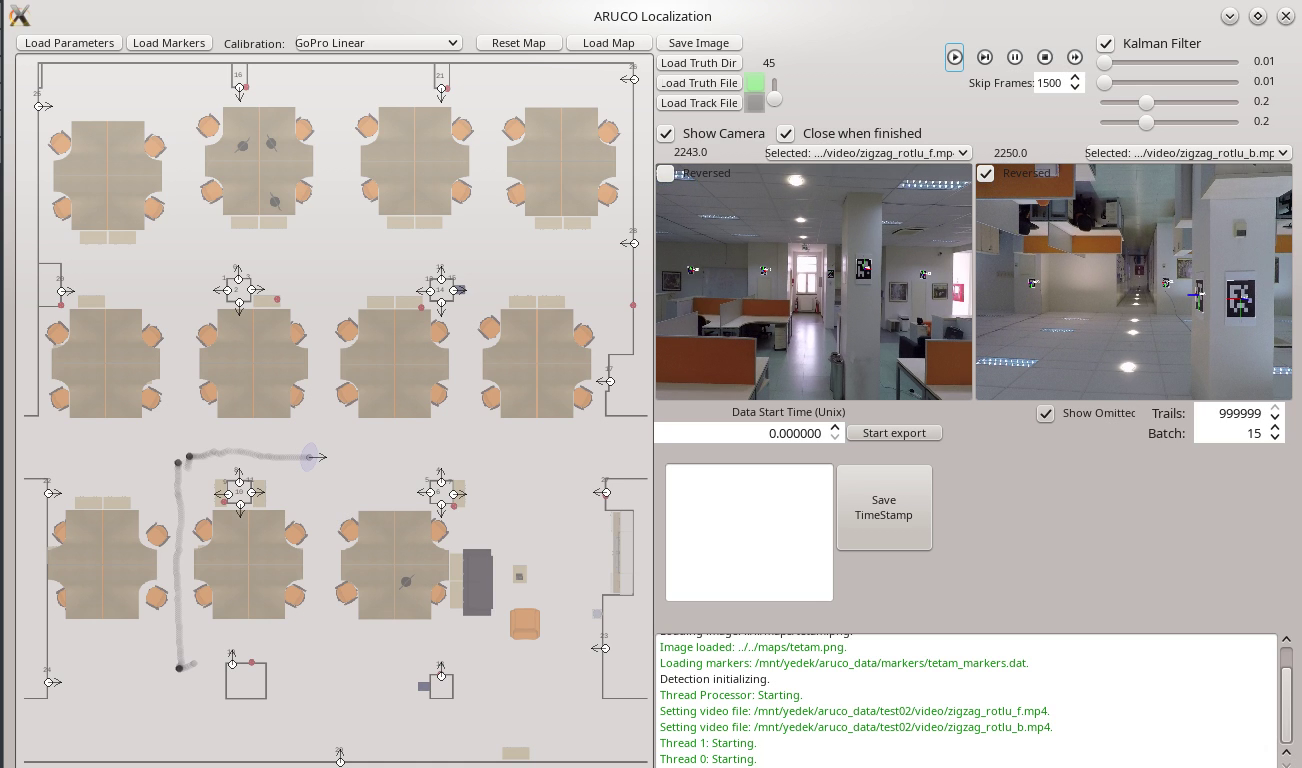}
\caption{AR localization toolkit for estimating, visualizing, synchronizing and logging data. \label{fig:ar_toolkit}}
\end{figure}

To ease the synchronization of video and RSSI data, we develop and utilize a custom toolkit written with PyQt5. A snapshot of the toolkit can be seen in \figurename{~\ref{fig:ar_toolkit}}. The toolkit is used for performing several tasks required for calibration, pose estimation and synchronization processes: such as displaying video frames from two cameras and navigate through the frames freely; detecting markers given the proper calibration parameters and displaying them in video frames; estimating raw positions given the markers; applying different elimination strategies; toggling Kalman filter and tuning its noise parameters on the fly; displaying raw, eliminated or smoothed poses on the map with their related markers; taking snapshots from the map (most of the figures including maps in this article are generated by this toolkit); marking the needed timestamps for synchronization; displaying logged or predetermined trajectories; and logging the obtained poses into data files.

\subsection{Data collection experiments}

\begin{figure}[!ht]
\centering
\subcaptionbox{Detected markers in forward and backward camera views (left) and corresponding raw pose estimations (right).\label{fig:marker_detection}}{
\includegraphics[width=.255\linewidth]{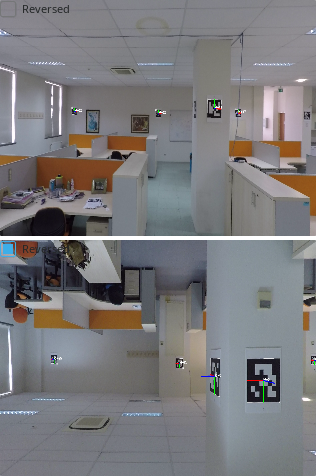}
\includegraphics[width=.34\linewidth]{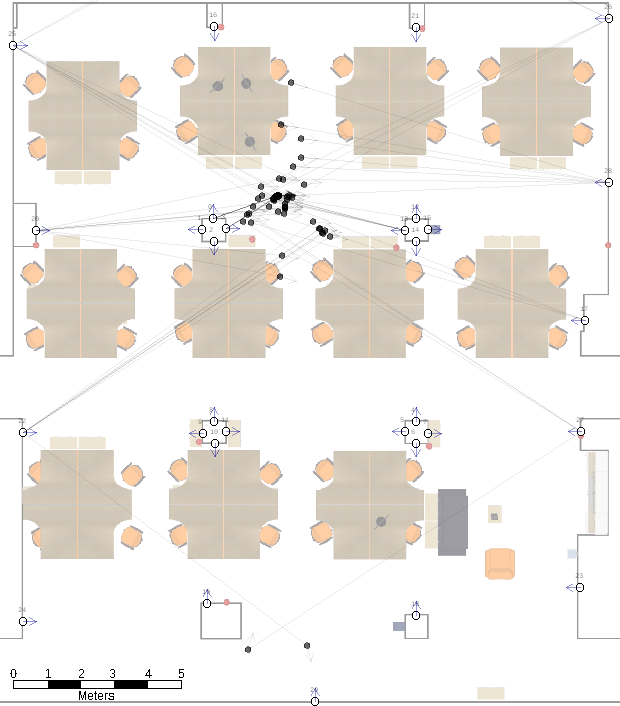}
}
\hfill
\subcaptionbox{Predetermined paths of the experiments.\label{fig:ideal_paths}}{
\includegraphics[width=.34\linewidth]{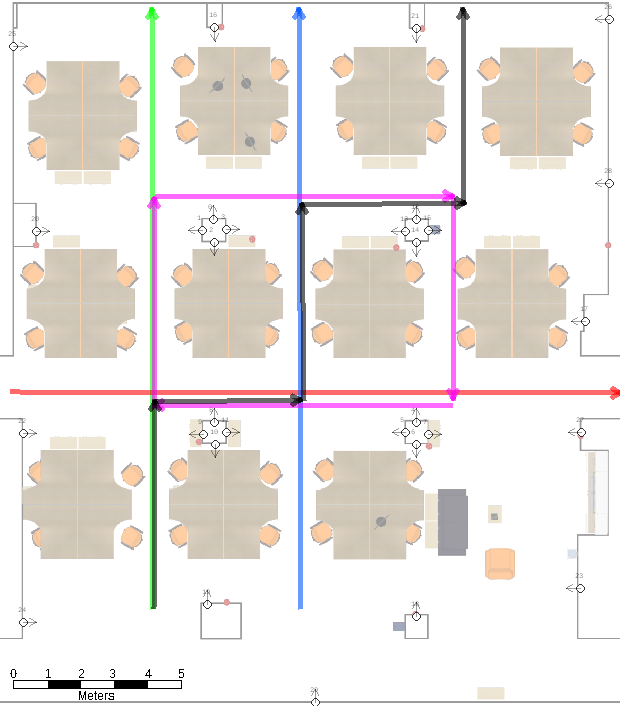}}
\caption{Position data collection and experiments}
\end{figure}

We collect two types of data simultaneously: timestamped RSSI data captured by BLE sensors, and position estimations through the videos captured by the cameras (see \figurename{~\ref{fig:marker_detection}}).

We prepare five experimental predetermined paths for testing the performance of our methodology: three distinct straight paths, a rectangular path and a zigzagging one. The camera-beacon bundle is driven along these paths trying to respect a linear and smooth motion, simulating human movement. The predetermined paths are also given in \figurename{~\ref{fig:ideal_paths}}. The data collection processes are varied by changing the pace on the straight paths and adding rotation on the rectangular and zigzagging paths, which makes a dataset of 15 trajectories.

\begin{figure}[!ht]
\centering
\includegraphics[width=.24\linewidth]{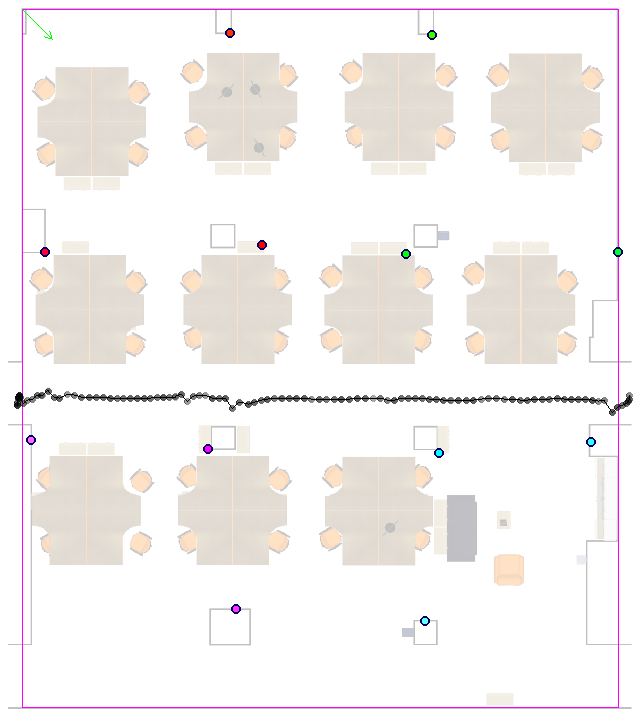}
\includegraphics[width=.24\linewidth]{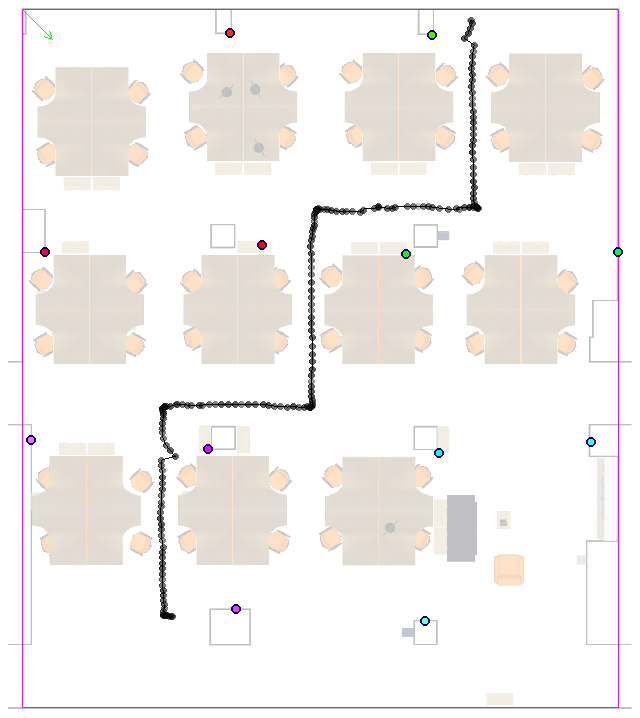}
\includegraphics[width=.24\linewidth]{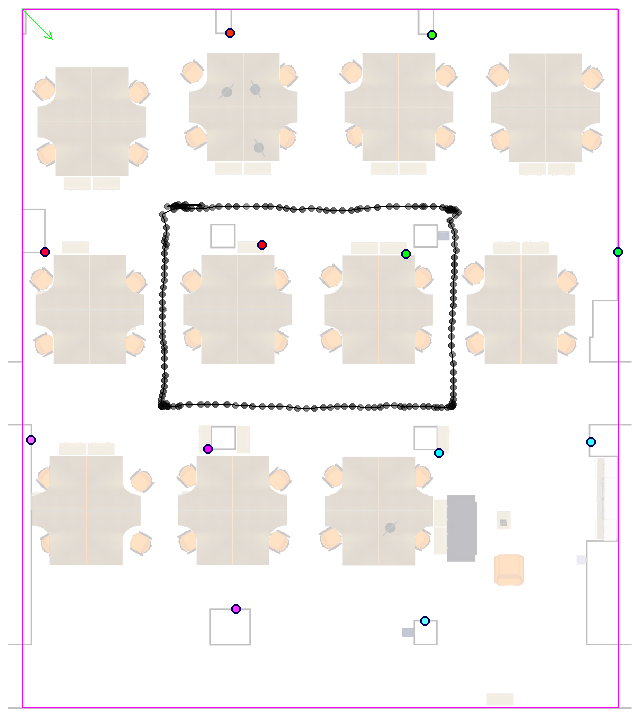}
\includegraphics[width=.24\linewidth]{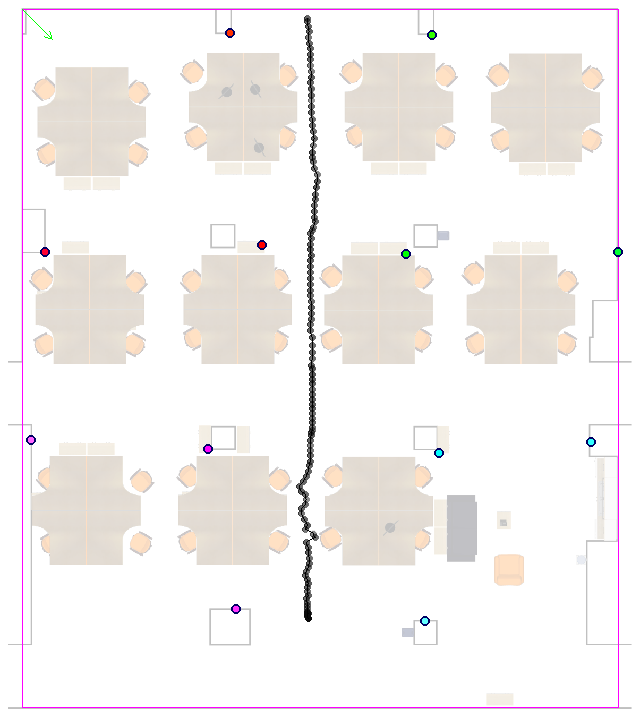}
\includegraphics[width=.24\linewidth]{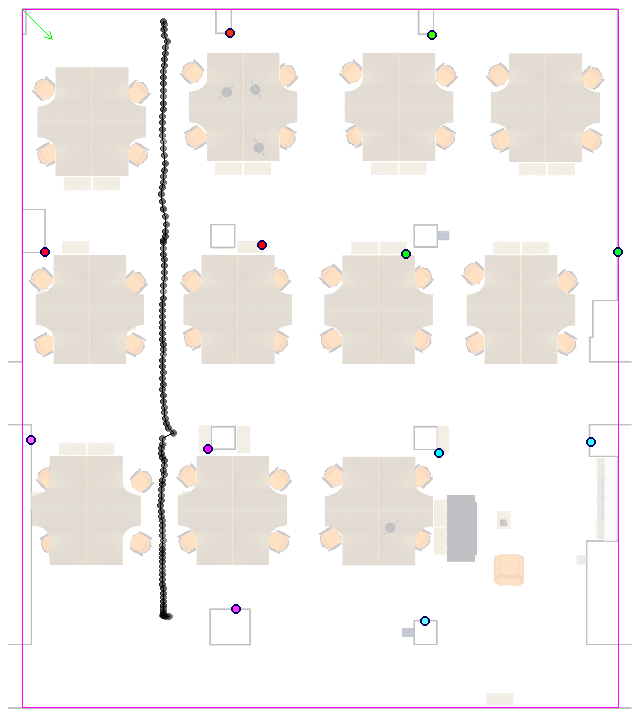}
\includegraphics[width=.24\linewidth]{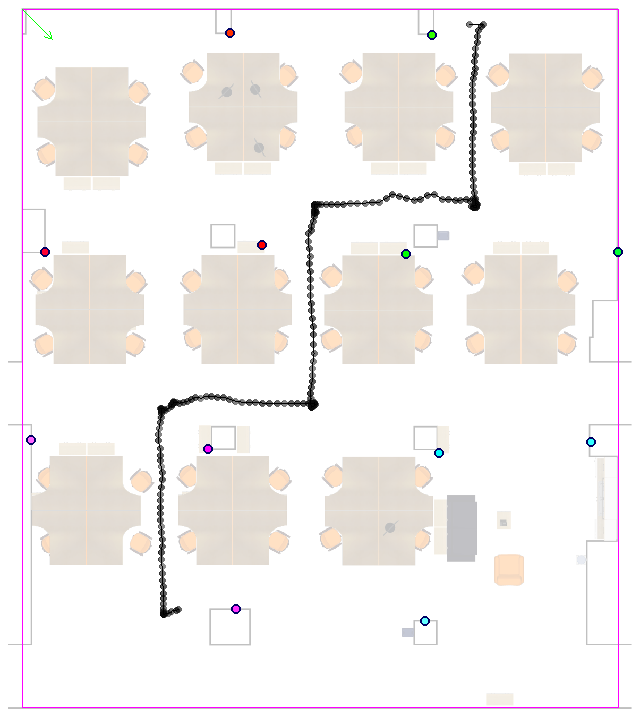}
\includegraphics[width=.24\linewidth]{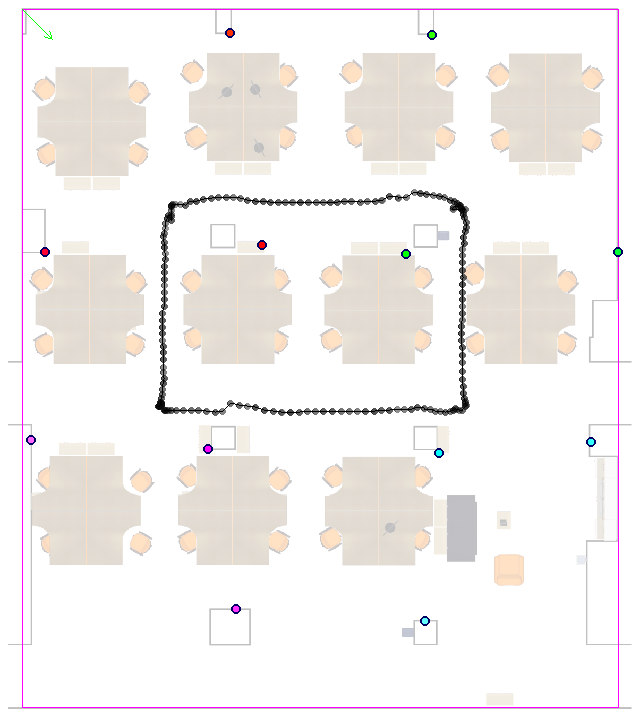}
\includegraphics[width=.24\linewidth]{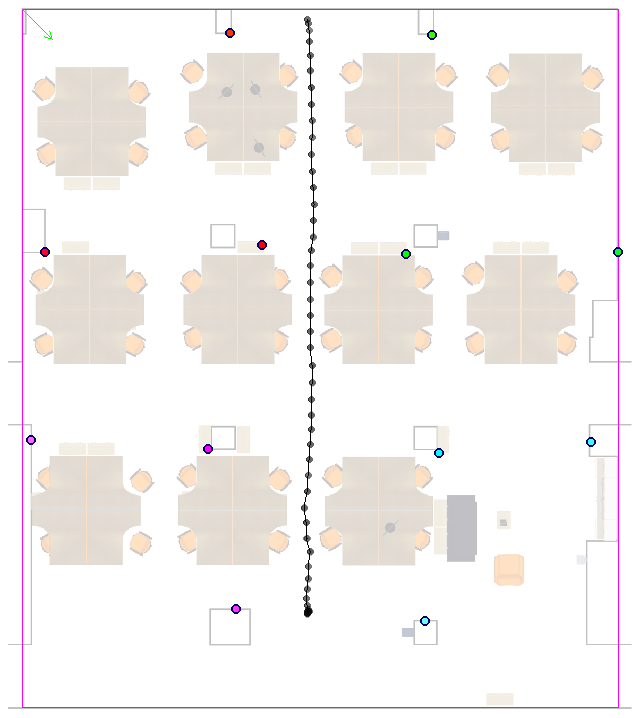}
\caption{Sample paths of collected data: Different straight paths, bendy paths with and without point rotations, and paths at different paces.\label{fig:trajectories}}
\end{figure}

The collected data are processed by varying parameters for the elimination strategies, the closest selection, \selection, and the outlier elimination, \outlier; and with the measurement error matrix of Kalman filter, $\obsN = q \identity_d$. We predefine the transition noise matrix as $\transN_t = 0.01 \identity_d$, which favors the pose estimations being closer to the previous ones. Some of the estimated paths are given in \figurename{~\ref{fig:trajectories}}.

There is a reason that we collect data from more complicated paths along with the pure straight paths. The rectangular paths investigate an expected phenomenon whether the beginning and the end of the trajectories point the same positions on the map, closing the loop. We observe that these same positions (and even orientations) are successfully captured. Moreover, we witness that even a point rotation may complicate the path on its own, because of the relative position of the camera-beacon bundle with respect to the rotation axis. However, we see that the selected parameters of the Kalman filter can compensate for this jumping behavior successfully. In our perspective, a complicated path refers to different things: this may be related to paths with turns and covering more space in the map, but we think that the pace makes things more complicated than choosing bendy paths, so we investigate this phenomenon and add the related results at the end of Section~\ref{sec:res}.

\subsection{Position annotated dataset}

The dataset provides the trajectory data files that include the position annotated RSSI data collected while navigating the special platform in the test area. There are 15 different trajectory files with varying lengths, forms and paces. Each sample in a file comes with a timestamp, a transmitter address, a sensor address, an RSSI reading, a ground truth position in 3D and its associated rotation matrix. We also supply the map of the test area, the files for the configuration of the sensors and various parameters that are required to convert the measurements in meters into pixels, and vice versa. For the completeness of an indoor localization project, we publish auxiliary data of raw fingerprints, their processed histograms, two sets of estimated probabilistic radio frequency maps in the grid form and sets of occupancy maps \citep{danis2021}.

\section{Results}
\label{sec:res}

We hypothesize that raw poses of the camera-beacon bundle display two types of faults, by visualizing raw poses given in \figurename{~\ref{fig:marker_detection}}: (i) they may be highly erroneous and positioned very far from the true position, because of highly incorrect and inverted axes of occluded or very distant markers (e.g. two points in the lower part of the map), or (ii) they may fluctuate around a true point because of pixel-wise problems with smaller errors (e.g. the ones around the dominant cluster). The former types of errors can be controlled by discarding the estimations due to the farthest markers, and then an outlier targeted pruning can be employed on the remaining poses for the latter cases.

\begin{figure}[!ht]
\centering
\includegraphics[width=.9\linewidth]{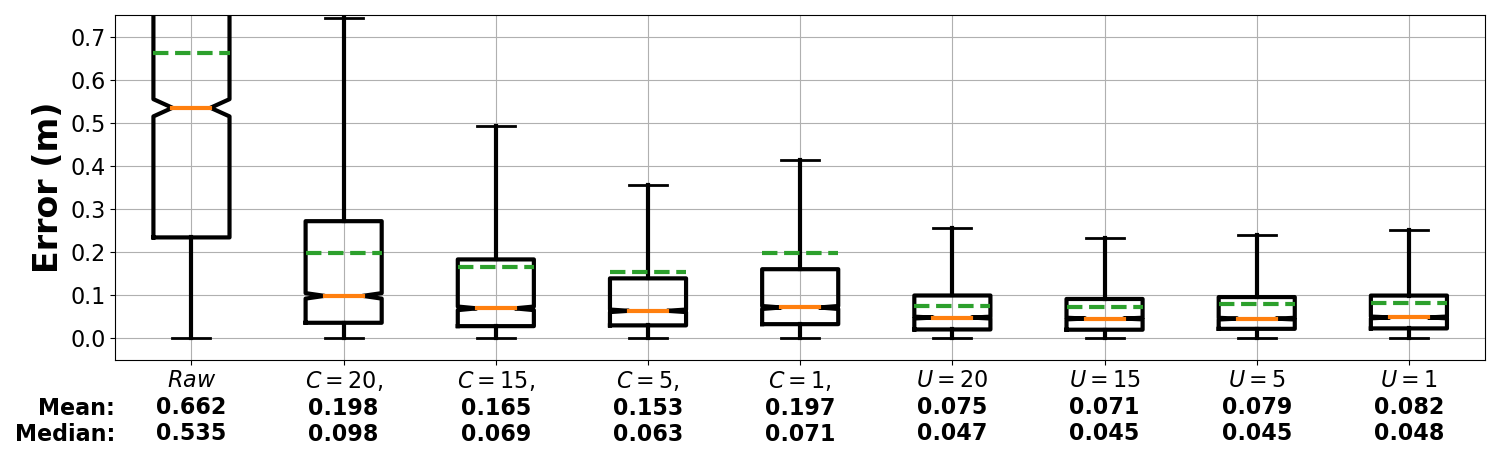}
\caption{Deviations from the predetermined linear paths of raw poses and isolated elimination strategies: \selection~for close marker selection and \outlier~for outlier elimination. \label{fig:isolated}}
\end{figure}

To evaluate the estimated trajectories, we report how the estimations deviate from the predetermined linear paths, knowing that data are collected by the experimenter that follows straight segments as given in \figurename{~\ref{fig:ideal_paths}}. We first compare the isolated versions of the two elimination strategies with the errors of raw poses (without elimination). According to \figurename{~\ref{fig:isolated}} and in accordance with our hypothesis, it is obvious that a pose elimination strategy is mandatory, and between these isolated elimination strategies, the outlier elimination strategy with $\outlier = 15 $ performs better than others, giving an error median of 0.045. Moreover, the error distributions for the selection strategy display undesired skewed distributions with high numbers of outliers. However, even though a pure close marker selection strategy performs significantly worse than a pure outlier elimination strategy, we still prefer to begin with a selection strategy because of its speed and ability to discard probable systematic outliers due to misestimated rotation axes.

\begin{figure}[!ht]
\centering
\includegraphics[width=.9\linewidth]{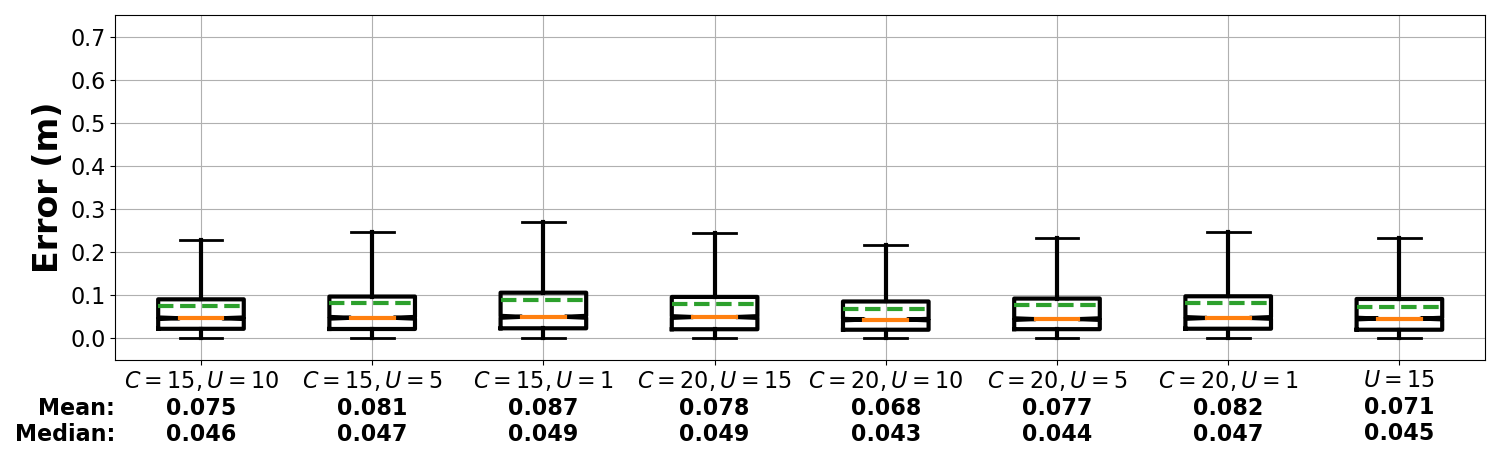}
\caption{Deviations from the predetermined linear paths of the elimination strategy combinations with different parameters: \selection~for close marker selection and \outlier~for outlier elimination.\label{fig:combined}}
\end{figure}

In \figurename{~\ref{fig:combined}}, we compare different combinations of the pose elimination strategies with the isolated selection strategy with $\outlier = 15$. Whereas there is not a significantly best pose elimination combination, a combination of a selection strategy with $\selection = 20$ and a succeeding outlier elimination with $\outlier = 10$ distinguishes among the experiments performed, with a median and mean deviations of 0.043 m and 0.068 m, respectively. Usage of a close marker selection is also preferable because a direct application of the greedy outlier elimination strategy is computationally intensive, so it would be also intuitive to precede it with the quicker close marker selection strategy, since the far markers tend to yield misleading estimations. Aligned with this preference, we select a combination of the pose elimination strategies with $\selection = 20$, $\outlier = 10$, which corresponds to quickly reducing the number of poses from about 100-150 down to 20 with the close marker selection strategy, and then applying the outlier elimination method on a set of 20 poses to obtain a final set of 10 poses.

\begin{figure}[!ht]
\centering
\includegraphics[width=.9\linewidth]{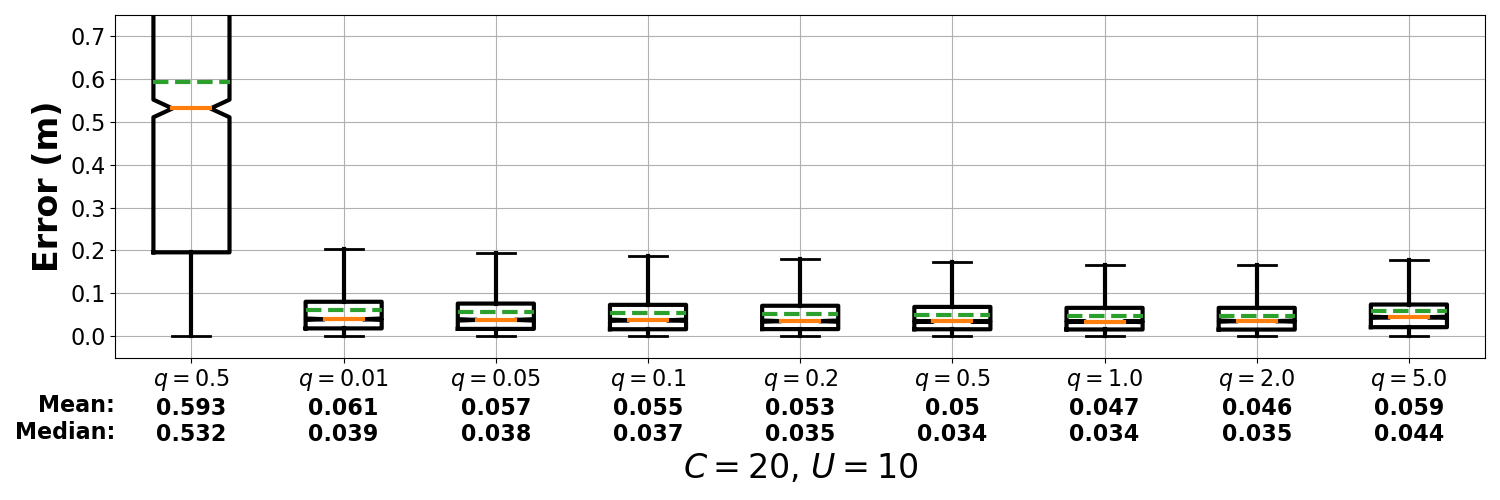}
\caption{Deviations from the predetermined linear path after the Kalman filter with different noise covariance multipliers, preceded by a combination of selection and outlier detection strategies. \label{fig:kalman_covariance}}
\end{figure}

On the selected pose estimations after the elimination techniques, we employ a Kalman filter to attain smoother trajectories. \figurename{~\ref{fig:kalman_covariance}} summarize the parameters for the noise covariance multiplier, $q$, and the deviations after employing the Kalman filter on the preprocessed pose estimations. We decide that the deviation medians can be reduced down to 0.034 meters and the means under 0.05 meters with a covariance matrix of $\obsN = 2.0 \identity_d$.

\begin{figure}[!ht]
\centering
\includegraphics[width=.9\linewidth]{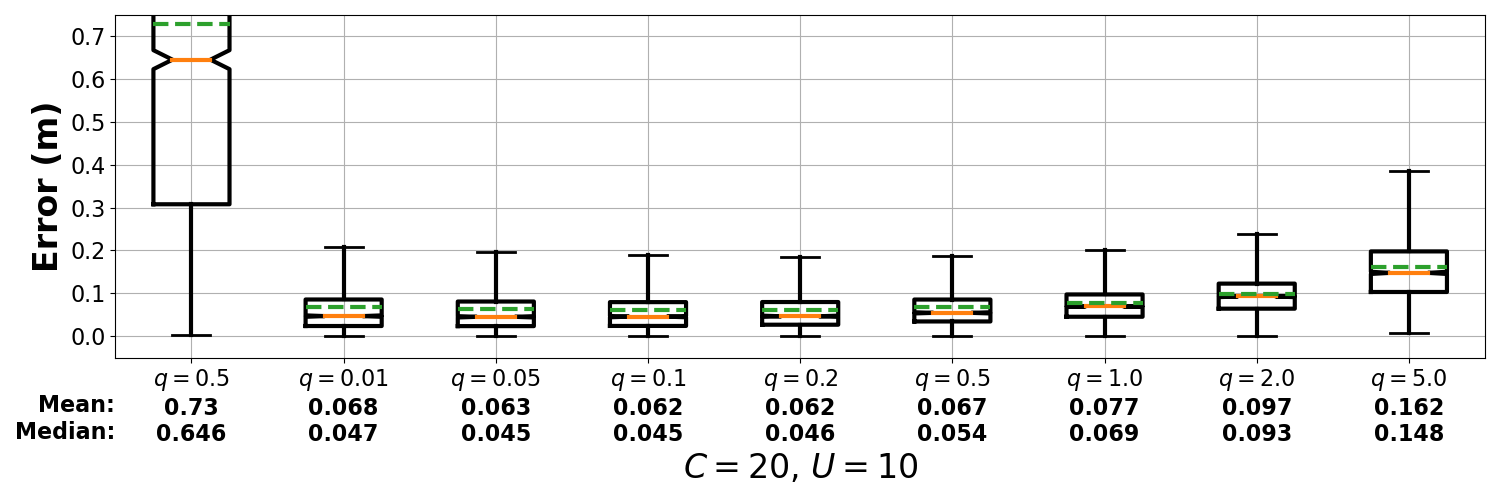}
\caption{Deviations from the predetermined linear path after the Kalman filter with different covariance multipliers and lag penalty application.\label{fig:kalman_penalty}}
\end{figure}

Smoothing with the Kalman filter misleads to inaccurate trajectories, because we discard time up to this point. We measure and compare the deviations of the estimations from the predefined ideal paths. Smoother paths yield lower deviations, however, when the Kalman filter is applied on the pose estimations, dragging is also possible if the noise covariance multiplier, $q$, of the Kalman filter is chosen to be too high. If this covariance is excessively high the estimated trajectory is surely smooth, yielding very low deviances, but fails to estimate the real positions with respect to time. This dragging can be evaluated by measuring the distance between a real known end point of a segment and the estimated points at the same timestamp. To avoid high dragging situations, we penalize the experiment by accumulating these drag distances as errors on the deviations, and we obtain \figurename{~\ref{fig:kalman_penalty}}. The parameters with $q > 0.2$ are highly penalized, meaning that estimated paths do not lead to accurate localization, and with $q < 0.1$, less smoothing is applied. We conclude that choosing $q \in [0.1, 0.2]$ for these experiments results with the best performances.

\begin{figure}[!ht]
     \centering
\subcaptionbox{\label{fig:results_raw}}{\includegraphics[width=.22\linewidth]{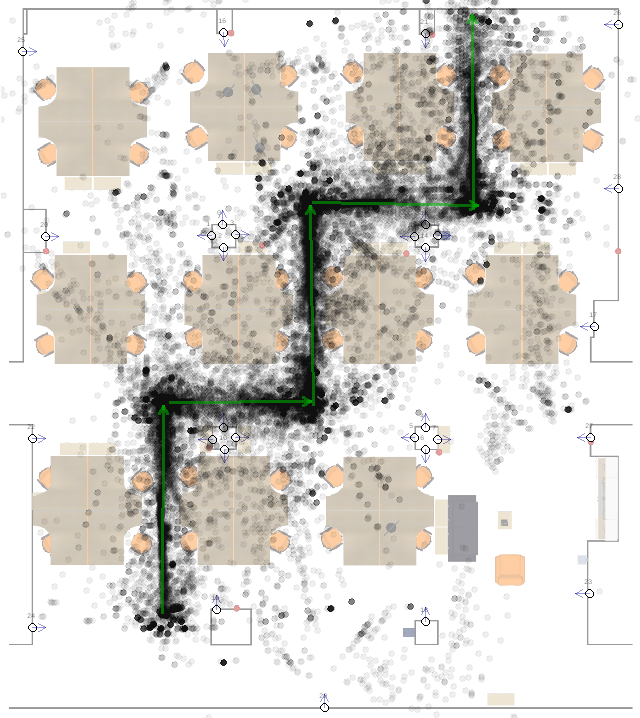}}
\subcaptionbox{\label{fig:results_kalman}}{\includegraphics[width=.22\linewidth]{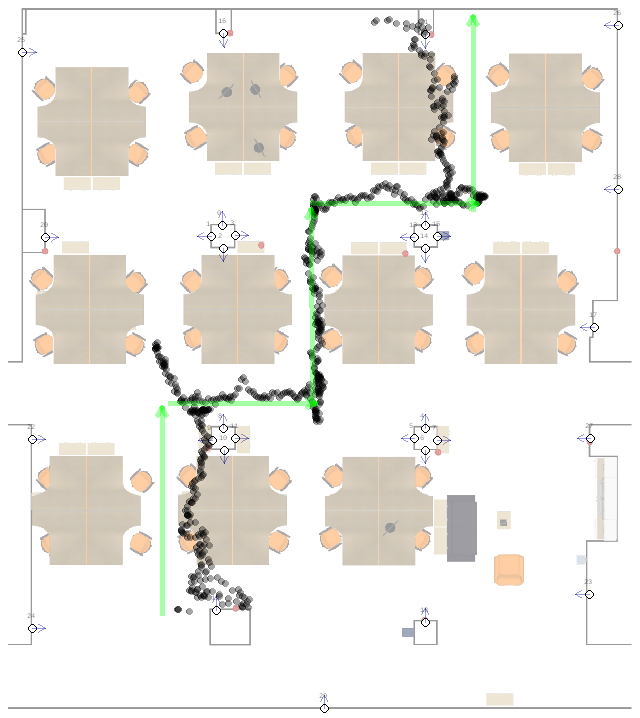}}
\subcaptionbox{\label{fig:results_elimination}}{\includegraphics[width=.22\linewidth]{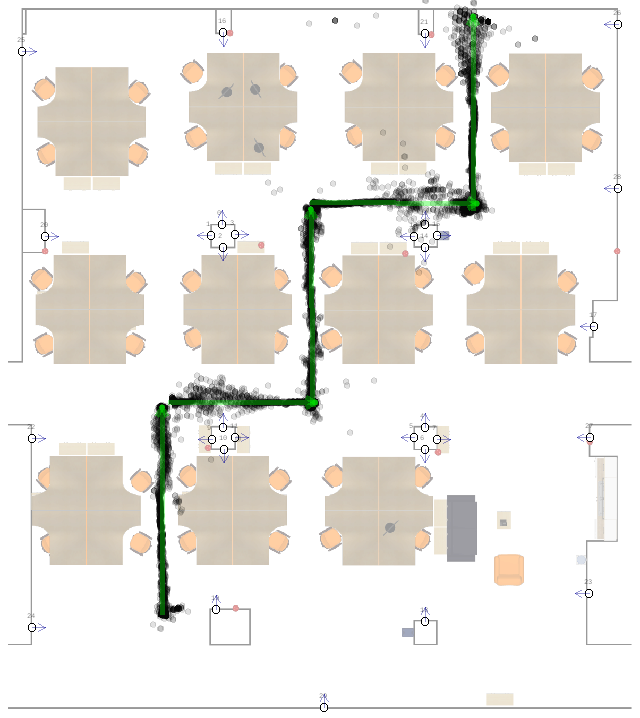}}
\subcaptionbox{\label{fig:results_full}}{\includegraphics[width=.22\linewidth]{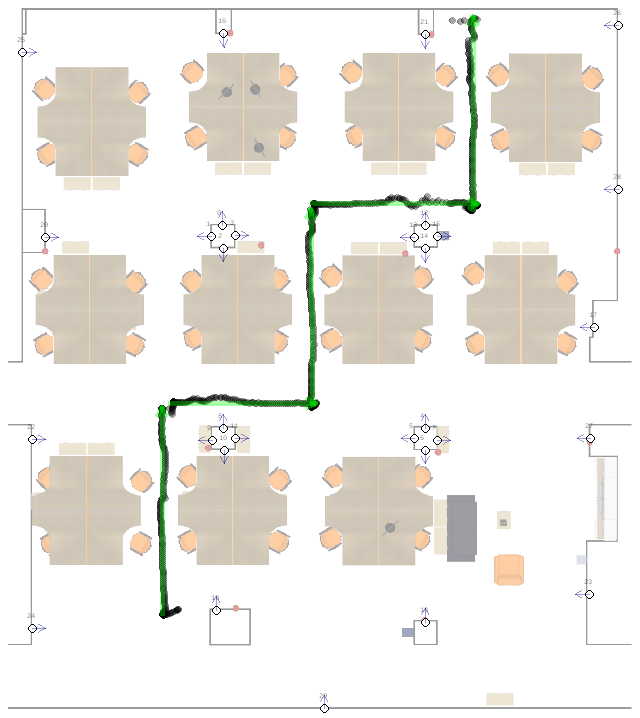}}
\caption{Experiments with different strategies with black dots denoting the estimated positions and green segments the predefined paths: (a) raw position estimations, (b) estimations after only the Kalman filter, (c) only elimination strategies, (d) and the Kalman filter preceded by elimination strategies.}
\label{fig:zigzag}
\end{figure}

One may propose to apply a Kalman filter directly on the raw position estimations (\figurename{~\ref{fig:results_raw}}) without any prior elimination strategy (see \figurename{~\ref{fig:results_kalman}}), however, as the raw poses are highly erroneous because of mainly misdetection of marker orientations, the Kalman filter is unable to cope with raw poses unaidedly. An elimination strategy is mandatory. The positions after the elimination strategies still draw fluctuating trajectories as seen in \figurename{~\ref{fig:results_elimination}}, but the selected position estimations are distributed around the true paths. These fluctuations may be eliminated with online filtering mechanisms. \figurename{~\ref{fig:results_full}} is the result of first applying the aforementioned elimination strategies on the positions estimated by the ArUco marker detection algorithm, and then smoothing the intermediate results using the Kalman filter with proper noise covariance parameters. We see that the green predetermined path can also be visually matched with the filtered positions.

\begin{figure}[!ht]
\centering
    \includegraphics[width=.8\linewidth]{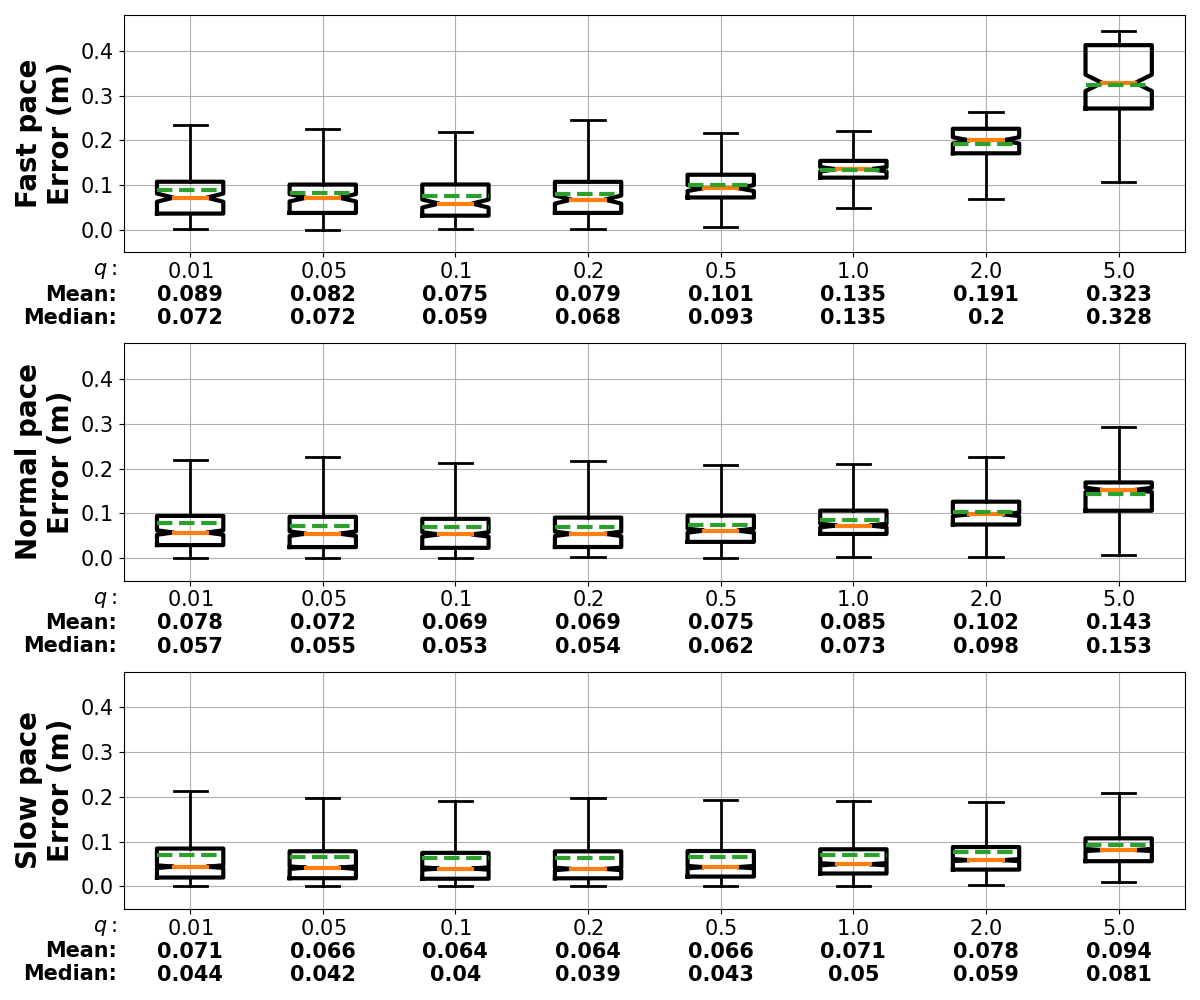}
\caption{Deviations from the predetermined linear path after employing the Kalman filter with different noise covariance multipliers and lag penalty application for the experiments at varying paces.}
    \label{fig:kalman_pace}
\end{figure}

We consider a final situation where Kalman parameters are also affected by the speed of the navigation. We also test the same experiment at three different paces. The results in \figurename{~\ref{fig:kalman_pace}} show that choosing $q = 0.2$ and $q = 0.1$ for slow navigations (at about 0.3 m/s) and for quicker navigations (at about 0.4 m/s) respectively will yield better performances.

\section{Conclusion}
\label{sec:conc}

This work proposes techniques for collecting high precision position data using AR markers and annotating BLE RSSI with precise positions in both time and space domain. We decorate a 364 square meter office area with AR markers and BLE sensing equipment to collect simultaneous video and RSSI data using a low cost and practical custom setup. The videos are used to estimate the noisy positions, which are then eliminated and filtered into smoother trajectories, highly correlated with the ideal paths. With the aforementioned processing and synchronization techniques, the RSSI data can be annotated by the precise position estimations to obtain position annotated RSSI datasets.

The results show that with AR markers we can achieve very high precision positioning (under 0.05 meters) in controlled indoor areas. This makes the proposed AR based positioning method far more accurate than the state-of-art WiFi or BLE based positioning. Of course, the aim is not to compare the positioning systems. Instead, the proposed setup and method can serve the indoor positioning domain as a benchmark tool to measure and verify other yet less precise localization performances in the indoor positioning domain. Moreover, we believe that there still exists some area to make the AR based positioning system more precise, especially with better illumination conditions, more rigorous measurements or larger and more markers.

In the literature, indoor positioning algorithms are evaluated mostly by manually measuring the predefined reference points. The lack of high frequency position annotations prevents the community from standardizing the evaluation procedures. The position estimations of the tracking algorithms are assessed using various distinct techniques. This need makes high frequency ground truth data very valuable. Besides, we bring forth not only a method for fine grained ground truth data collection, but also a method for collecting data practically. Considering the lack of high precision data sets in the indoor positioning community, we make the data publicly available for the researchers who want to test their algorithms based on RSSI data.

We regard the AR based localization as a one-time prior assistive tool for the calibration of a wireless indoor positioning system. It is intended for collecting position annotated data which can later be used for parameter estimation of the tracking algorithms, fingerprinting and more importantly evaluation of the whole system in terms of position estimation accuracy.

Whereas, in this work, the data are shown to be processed and recorded by a single node, the infrastructure and the dataset allow the researchers to approach the wireless based indoor tracking problem through a distributed and decentralized manner. As an example, as a future work, it can be considered to handle the problem in a pure decentralized way, where each computer or sensor node is to estimate the position of an emitter and share its belief with the neighboring processors. This approach will make the setup easily scalable to any area.

The main focus of this work is to propose a method for collecting position annotated wireless data. Nevertheless the most prominent deficiency of the output is that the dataset is not constructed to reflect the human body influence on the wireless signals, even though there is a person that drives the navigable platform. As wireless based positioning systems are usually considered to be human or living things centered, the lack of this influence in the dataset may be considered important. Moreover, the dataset cannot be projected to the true nature of the crowded environments in terms of the signal quality. However, for the purposes of tracking non-living things or living things in sparse environments, the dataset still reflects the true nature and is valid to become a benchmark testing of wireless based positioning and tracking systems. Besides, the method itself and the system architecture are not prone to human body influence as they are already designed to be redundant and scalable.

Collecting wireless data using the current setup and constructing a new dataset influenced by the human body are left as future work highly due to the current pandemic situation. The future data set should include varying versions of the same experiment where the object to be tracked is held, pocketed or worn as an accessory. Moreover varying the movement and the number of people is another issue that will present different natural conditions.

As more future work, with proper tunings, this method can be used for automatic, online and fast fingerprinting for wireless signal based localization, which is substantially a challenging task for the researchers working in the domain. With the markers installed in the area, the camera-beacon bundle may be navigated to collect data from any point, and these data can be processed to form the fingerprints fast and easily. Moreover, a self navigating robot can be used for real automatization of fingerprinting. Furthermore, the technique also permits collecting data in three dimensions, which will prove to be essential for indoor localization of robots, specifically drones.


\section*{Acknowledgment}

This work is supported by the Turkish Directorate of Strategy and Budget under the TAM Project number 2007K12-873. The authors would like to thank Sergio Bromberg for his contribution of the back to back GoPro camera mount 3D model to Thingiverse and PoiLabs for supplying POI Beacons for this work.




\section*{\refname}
\bibliographystyle{elsarticle-num}
\bibliography{ground_truth_with_markers}

\end{document}